\documentclass[10pt,twocolumn,letterpaper]{article}

\usepackage{cvpr}
\usepackage{pifont}
\usepackage[utf8]{inputenc}
\usepackage{times}
\usepackage{epsfig}
\usepackage{graphicx}
\usepackage{amsmath}
\usepackage{amssymb}
\usepackage{algorithm}
\usepackage{algpseudocode}
\usepackage{float}
\usepackage{color}
\usepackage{colortbl}
\usepackage{url}
\usepackage{breqn}
\usepackage{rotating}
\usepackage{times}
\usepackage{epsfig}
\usepackage{enumitem}
\usepackage{makecell}
\usepackage{lineno}
\usepackage{tabularx}
\usepackage{xcolor}
\usepackage{multirow}
\usepackage{graphicx}
\usepackage{hhline}
\usepackage{amsmath}
\usepackage{amssymb}
\usepackage{amsfonts}
\usepackage{empheq}
\usepackage{float}
\usepackage{framed, color}
\usepackage{xcolor}
\usepackage{graphics}
\usepackage{graphicx}
\usepackage[]{graphicx} % more modern
\usepackage{epsfig} % less modern
\usepackage{color}
\usepackage{filecontents}
\usepackage{subfigure}
\usepackage{multirow}
\usepackage{filecontents}
\usepackage{verbatim} % begin{comment} end{comment}
\usepackage{tabularx}
\usepackage{lineno}
\usepackage{setspace}
\usepackage{textcomp,booktabs}
\usepackage{caption}
\usepackage{stmaryrd}
\usepackage[mathscr]{eucal}
\usepackage{lineno}
\usepackage{subfigure}
\usepackage[export]{adjustbox}
\usepackage{epsfig}

\usepackage{times}
\usepackage{epsfig}
\usepackage{graphicx}
\usepackage{amsmath}
\usepackage{amssymb}

\def\0{{\bf 0}}
\def\1{{\bf 1}}

\definecolor{hyperlinkTextType}{rgb}{0.8, 0.0, 0.0}

\definecolor{red}{rgb}{0.95,0.4,0.4}
\definecolor{purered}{rgb}{1,0,0}
\definecolor{blue}{rgb}{0.4,0.4,0.95}
\definecolor{darkblue}{rgb}{0,0,0.8}
\definecolor{darkred}{rgb}{1,0,0}
\definecolor{darkgreen}{rgb}{0,0.5,0}
\definecolor{grey}{rgb}{0.6,0.6,0.6}
\definecolor{col1}{RGB}{232, 161, 148}
\definecolor{col2}{RGB}{148, 187, 232}
\definecolor{lightgrey}{rgb}{0.85,0.85,0.85}
\definecolor{lightlightgrey}{rgb}{0.9,0.9,0.9}
\definecolor{verylightBG}{rgb}{0.9,0.99,0.99}
\definecolor{darkgreen}{rgb}{0.3, 0.75, 0.3}

\graphicspath{{./figure/}}

\newcommand\solved[1]{\textcolor{black}{#1}}

\newcommand\note[1]{\textcolor{red}{#1}}

\usepackage[pagebackref=true,breaklinks=true,colorlinks,bookmarks=false]{hyperref}
\cvprfinalcopy % *** Uncomment this line for the final submission

\begin{document}

\title{\emph{OpenGAN}: Open-Set Recognition via Open Data Generation}

\author{Shu Kong\textsuperscript{$\ast$}, \ \   Deva Ramanan\textsuperscript{$\ast$,$\dag$} \\
\textsuperscript{$\ast$}Carnegie Mellon University   \ \ \ \ \   \textsuperscript{$\dag$}Argo AI \\
{\small \{\tt shuk, deva\}@andrew.cmu.edu}
\\
\ [\href{https://github.com/aimerykong/OpenGAN}{{\bf Github Repository}}]
}

\maketitle

\begin{abstract}
Real-world machine learning systems need to analyze test data that may differ from training data. In K-way classification, this is crisply formulated as open-set recognition, core to which is the ability to discriminate open-set data outside the K closed-set classes. Two conceptually elegant ideas for open-set discrimination are: 1) discriminatively learning an open-vs-closed binary discriminator by exploiting  some outlier data as the open-set, and 2) unsupervised learning the closed-set data distribution with a GAN,  using its discriminator as the open-set likelihood function. However, the former generalizes poorly to diverse open test data due to overfitting to the training outliers, which are unlikely to exhaustively span the open-world. The latter does not work well, presumably due to the instable training of GANs. Motivated by the above, we propose OpenGAN, which addresses the limitation of each approach by combining them with several technical insights. First, we show that a carefully selected GAN-discriminator on some real outlier data already achieves the state-of-the-art. Second, we augment the available set of real open training examples with adversarially synthesized ``fake'' data. 
Third and most importantly, we build the discriminator over the features computed by the closed-world K-way networks. This allows OpenGAN to be implemented via a lightweight discriminator head built on top of an existing K-way network.
Extensive experiments show that OpenGAN significantly outperforms prior open-set methods.
\end{abstract}

%%%%%%%%% BODY TEXT
\section{Introduction}
\label{sec:intro}

\begin{figure}[t]
\centering
\includegraphics[width=0.99\linewidth]{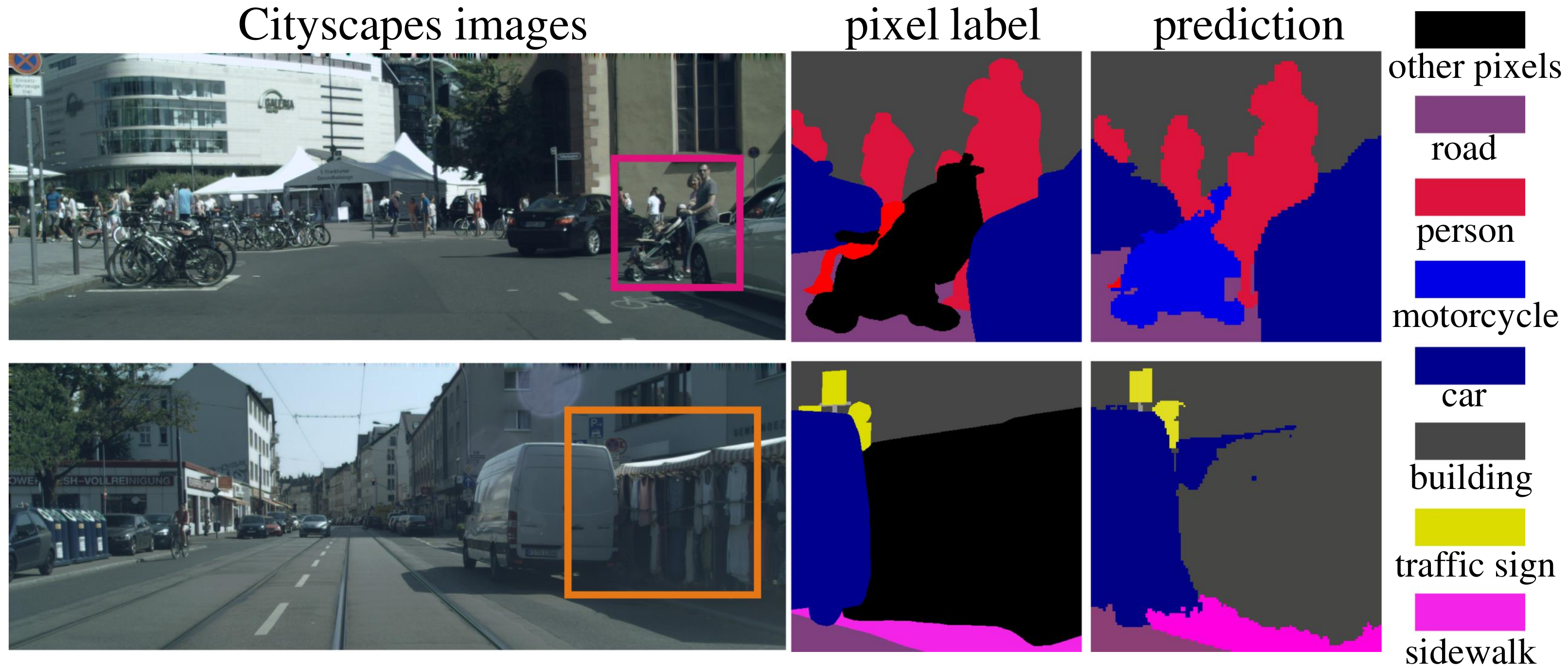}
\vspace{-3mm}
\caption{\small
We motivate open-set recognition with safety concerns in autonomous vehicles (AVs).  
Contemporary benchmarks such as Cityscapes~\cite{cordts2016cityscapes} focus on $K$ classes of interest for evaluation, ignoring a sizeable set of ``other'' pixels that include vulnerable objects like wheelchairs and strollers (upper row).
As a result, most state-of-the-art segmentators~\cite{WangSCJDZLMTWLX19} also ignore these pixels during training, resulting in a \textcolor{red}{stroller} mislabeled as a ``motorcycle'' (top) and a \textcolor{orange}{street-market} mislabeled as a ``building''. Such misclassifications may be critical for AVs because these objects may require different plans for obstacle avoidance (e.g., ``yield'' or ``slow-down''). Fig.~\ref{fig:qualitativeDemo} shows our approach, which explicitly augments state-of-the-art segmentors with open-set reasoning.
}
\vspace{-5mm}
\label{fig:openset-pixel}
\end{figure}

Machine learning systems that operate in the real open-world invariably encounter test-time data that is unlike training examples,  such as anomalies or rare objects that were insufficiently (or never) observed during training. Fig.~\ref{fig:openset-pixel} illustrates two cases in which a state-of-the-art semantic segmentation network
misclassifies a ``stroller''/ ``street-market'' ---  a rare occurrence in either training or testing --- as a ``motorcycle''/``building''.
This failure could be catastrophic for an autonomous vehicle.

Addressing the open-world has been explored through anomaly detection~\cite{zong2018deep, hendrycks2018deep} and out-of-distribution  detection~\cite{liang2017enhancing}.
In $K$-way classification, this task can be crisply formulated as open-set recognition,
which requires discriminating open-set data that belongs to a ($K$+1)$^{th}$ ``other'' class, outside the $K$ closed-set classes~\cite{scheirer2012toward}.
Typically, open-set discrimination assumes no examples from the ``other'' class available during training~\cite{bendale2016towards, Yoshihashi19, OzaP19}. In this setup, one elegant approach is to learn the closed-set data distribution with a GAN, making use of the GAN-discriminator as the open-set likelihood function (Fig.~\ref{fig:flowchart}a)~\cite{schlegl2017unsupervised, sabokrou2018adversarially, pidhorskyi2018generative, zenati2018adversarially, liu2019generative}. However, this does not work well due to instable training of GANs.
Recent work has shown that outlier exposure (Fig.~\ref{fig:flowchart}b), or the ability to train on {\em some} outlier data as open-training examples, can work surprisingly well via the training of a simple open-vs-closed binary discriminator~\cite{dhamija2018reducing, hendrycks2018deep}. 
However, such discriminators fail to generalize to diverse open-set data~\cite{ShafaeiSL19bmvc} because they overfit to the available set of training outliers, which are often biased and fail to exhaustively span the open-world.
Motivated by above, we introduce {\bf OpenGAN}, a simple approach that dramatically improves open-set classification accuracy by incorporating several key insights.
First, we show that using outlier data as a valset to select the ``right'' GAN-discriminator \emph{does} achieve the state-of-the-art on open-set discrimination.
Second, with outlier exposure, we augment the available set of open-training data by adversarially generating \emph{fake} open examples that fool the binary discriminator (Fig.~\ref{fig:flowchart}c).
Third and most importantly, rather than defining discriminators on pixels, we define them on off-the-shelf (OTS) features computed by the closed-world $K$-way classification network (Fig.~\ref{fig:flowchart}d).
We find such discriminators generalize much better.

Our formulation differs in three ways from other open-set approaches that employ GANs.
(1) Our goal is \emph{not} to generate realistic pixel images, but rather to learn a robust open-vs-closed discriminator that naturally serves as an open-set likelihood function. 
Because of this, our approach might be better characterized as a discriminative adversarial network! (2) We train the discriminator with both \emph{fake} data (synthesized from the generator) and {\em real} open-training examples (cf. outlier exposure~\cite{hendrycks2018deep}).
(3) We train GANs on OTS features rather than RGB pixels. We show that OpenGAN significantly outperforms prior work for open-set recognition across a variety of tasks including image classification and pixel segmentation.
Moreover, we demonstrate that our technical insights improve the accuracy of other GAN-based open-set methods: training them on OTS features and selecting their discriminators via validation as the open-set likelihood function.

\section{Related Work }
\label{sec:background}

{\bf Open-Set Recognition}. There are multiple lines of work addressing open-set discrimination, such as anomaly detection~\cite{chandola2009anomaly, liang2017enhancing, zong2018deep}, outlier detection~\cite{sabokrou2018adversarially, pidhorskyi2018generative}, and open-set recognition~\cite{scheirer2012toward, geng2020recent}.
The typical setup for these problems assumes that one does not have access to training examples of open-set data.
As a result, many approaches first train a closed-world $K$-way classification network on the closed-set~\cite{hendrycks2016baseline, bendale2016towards} and then exploit the trained network for open-set discrimination~\cite{scheirer2012toward, lee2018simple, OzaP19}.
Some others train ``ground-up'' models for both closed-world $K$-way classification and open-set discrimination by synthesizing fake open data during training, oftentimes sacrificing the classification accuracy on the closed-set~\cite{ge2017generative, neal2018open, Yoshihashi19, sun2020conditional}. 
To recognize open-set examples, they resort to post-hoc functions like density estimation~\cite{zong2018deep, zhang2020hybrid}, % , anonymous2021empirical
uncertainty modeling~\cite{gal2016dropout, liang2017enhancing}, and input image reconstruction~\cite{pidhorskyi2018generative, gong2019memorizing, dehaene2020iterative, sun2020conditional}.
We also explore open-set recognition through $K$-way classification networks, but we show OpenGAN, a simple and direct method of training an open-vs-closed classifier on adverserial data, performs significantly better than prior work.

{\bf Open-Set Recognition with GANs}. 
As GANs can learn data distributions~\cite{goodfellow2014generative}, conceptually, a GAN-discriminator trained on the closed-set naturally serves as an open-set likelihood function. However, this does not work well~\cite{schlegl2017unsupervised, sabokrou2018adversarially, pidhorskyi2018generative, zenati2018adversarially, liu2019generative}, presumably due to instable training of GANs. As a result, previous GAN-based methods focus on 1) generating fake open-set data to augment the training set, and 2) relying on the reconstruction error for open-set recognition~\cite{xia2015learning, schlegl2017unsupervised, sabokrou2018adversarially, akcay2018ganomaly, deecke2018image}. 
With OpenGAN, we show that GAN-discriminator can achieve the state-of-the-art for open-set discrimination \emph{once} we perform model selection on a valset of outlier examples. Therefore, unlike prior approaches, OpenGAN 
directly uses the discriminator as the open-set likelihood function.
Moreover, our final version of OpenGAN generates features rather than pixel images.

{\bf Open-Set Recognition with Outlier Exposure}.
\cite{dhamija2018reducing, hendrycks2018deep, ruff2019deep} reformulate the problem with the concept of  ``outlier exposure" which allows methods to access {\em some} outlier data as open-training examples. In this setting, simply training a binary open-vs-closed classifier works surprisingly well.
However, such classifiers easily overfit to the available set of open-training data and generalize poorly, e.g., in a ``cross-dataset" setting where open-set testing data differs from open-training data~\cite{ShafaeiSL19bmvc}.
It appears fundamentally challenging to collect outlier data to curate an exhaustive training set of open-set examples. 
Our approach, OpenGAN,
attempts to address this issue by augmenting the training set with adversarial {\em fake} open-training examples.

\section{OpenGAN for Open-Set Recognition}
\label{sec:discriminator for open-set}
Generally, solutions to open-set recognition contain two steps: (1) open-set discrimination that classifies testing examples into closed and open sets based on the open-set likelihoods, and (2) $K$-way classification on closed-set from step (1)~\cite{scheirer2012toward, bendale2015towards, OzaP19}.
The core problem to open-set recognition is the first step, i.e., open-set discrimination. Typically, open-set discrimination assumes that open-set examples are not available during training~\cite{neal2018open, OzaP19}.
However, \cite{dhamija2018reducing, hendrycks2018deep} demonstrate that outlier exposure, or the ability to train on {\em some} outlier examples, can greatly improve open-set discrimination via the training of a simple open-vs-closed binary classifier (Fig.~\ref{fig:flowchart}b).
Because it is challenging to construct a training set that exhaustively spans the open-world, such a classifier may overfit to the outlier data and not sufficiently generalize~\cite{ShafaeiSL19bmvc}.
We demonstrate that OpenGAN alleviates this challenge by generating fake open-set training examples using a generator that is adversarially trained to fool the classifier. 
Importantly, with model selection on a valset, OpenGAN is also effective under the classic setup which assumes no availability of open-training data.

\subsection{Methodology}
\label{ssec:opengan_method}

Let $x$ be a data example, which can be an RGB image or its feature representation. We will show that using the latter performs better.
Let ${\cal D}_{closed}(x)$ be the closed-world distribution over $x$ --- that is, closed-set data from the $K$ closed-set classes. Let ${\cal D}_{open}(x)$ be the open-set data distribution of examples which  do not belong to the closed-set.

{\bf Binary Classifier.}  We train a binary classifier $D$ from both closed- and open-set data:
\begin{equation}\small
\max_{D}
{\mathbb E}_{x\sim {\cal D}_{closed}} \big[ \log D(x)\big] +
\lambda_o \cdot {\mathbb E}_{x \sim {\cal D}_{open}} \big[ \log \big(1- D(x) \big) \big]
\nonumber
%\label{eq:binaryCls}
\vspace{-0.4mm}
\end{equation}
where $D(x)=p(y\text{=}{\text{``closed-set''}} \vert x)$.
Intuitively, we tune $\lambda_o$ to balance the closed- and open-set training examples. 
This simple method is effective when the open-training examples are sufficiently representative of testing-time open-set data~\cite{hendrycks2018deep},
but underperforms when they fail to span the open-world~\cite{ShafaeiSL19bmvc}.

{\bf Synthetic Open Data.} 
One solution to the above is to exploit synthetic training data, which might improve the generalizability of classifier $D$. Assume we have a generator network $G(z)$ that produces synthetic images given (Gaussian normal) random noise inputs $z \sim {\cal N}$. We can naively add them to the pool of negative or open-set examples that $D$ should not fire on. But these synthetic images might be too easy for $D$ to categorize as open-set data~\cite{nalisnick2018deep, chen2020learning}. A natural solution is to adversarially train the generator $G$ to produce difficult examples that fool the classifier $D$ using a GAN loss:
\begin{equation}\small
 \min_{G} {\mathbb E}_{z\sim  {\cal N}} \Big[ \log \big(1- D(G(z)) \big) \Big]
\label{eq:GAN_loss}
\vspace{-0.5mm}
\end{equation}
Because a \emph{perfectly} trained generator $G$ would generate realistic closed-set images, eventually making the discriminator $D$ inapplicable for open-set discrimination. 
We find that the following two techniques easily resolve this issue.

{\bf OpenGAN} trains with both the \emph{real} open\&closed-set data and the \emph{fake} open-data into a single (GAN-like) minimax optimization over $D$ and $G$:
\begin{equation}
\small
\begin{split}
%\displaystyle
\max_{D}\min_{G} &
{\mathbb E}_{x\sim {\cal D}_{closed}} \big[ \log D(x)\big] \\
& + \lambda_o \cdot {\mathbb E}_{\bar x \sim {\cal D}_{open}} \big[ \log \big(1- D(\bar x) \big) \big] \\
& + \lambda_G \cdot\mathbb{E}_{z\sim  {\cal N}} \big[ \log \big(1- D(G(z)) \big) \big]
% \nonumber
\end{split}
\label{eq:OpenGAN}
\end{equation}
where $\lambda_G$ controls the contribution of generated fake open-data by $G$.
When there are no open training examples (i.e., $\lambda_o$=$0$), the above minimax optimization can still train a discriminator $D$ for open-set classification.
In this case, training an OpenGAN is equivalent to training a normal GAN and using its discriminator as the open-set likelihood function. 
While past work suggests that GAN-discriminators do not work well as open-set likelihood functions, we show they \emph{do} achieve the state-of-the-art \emph{once} selected with a valset (detailed below). To distinguish our contribution on the crucial step of model selection via validation, we call this method OpenGAN-0.

{\bf Open Validation.}
Model selection is challenging for GANs.
Typically, one resorts to visual inspection of generated images from different model checkpoints to select the generator $G$~\cite{goodfellow2014generative}.
In our case, we must carefully select the discriminator $D$.
We experimented with many approaches such as using the last model checkpoint or selecting the one with minimum training error,
but neither works, because adversarial training will eventually lead to a discriminator $D$ that is incapable of discriminating closed-set data and fake open-set data generated by $G$ (details in the supplemental).
We find it crucial to use a validation set of real outlier data to select $D$, when $D$ achieves the best open-vs-closed classification accuracy on the valset.
We find the performance to be quite robust to the val-set of outlier examples, even when they are drawn from a different distribution from those encountered at test-time (Table~\ref{tab:heterogeneous-open-image} and \ref{tab:generalizability}).

\subsection{Comparison to Prior GAN Methods}
\label{sec:discussion}

We compare OpenGAN to numerous prior art that use GANs for open-set discrimination.

{\bf Discriminator vs. Generator}.
GANs mostly aim at generating realistic
images~\cite{arjovsky2017wasserstein, brock2018large}.
As a result, prior work in  open-set recognition has focused on using GANs to generate realistic open-training images~\cite{ge2017generative, lee2017training, neal2018open}. These additional images are used to augment the training set for learning an open-set model, which oftentimes is designed for both the closed-world $K$-way classification and open-set discrimination~\cite{ge2017generative, lee2017training, neal2018open}.
In our case, we do not learn a separate open-set model
but directly use the already-trained discriminator {\em as} the open-set likelihood function.

{\bf Features vs. Pixels}.
GANs are typically used to generate realistic pixel images.
As a result, many GAN-based open-set methods focus on generating realistic images to augment the closed-set training data~\cite{pidhorskyi2018generative, zenati2018adversarially, hendrycks2018deep}.
However, generating high-dimensional realistic images is challenging per se~\cite{arjovsky2017wasserstein, brock2018large} and may not be necessary to open-set recognition~\cite{pidhorskyi2018generative}.
As such, we build GANs over OTS feature embeddings learned by the closed-world $K$-way classification networks, e.g., over pre-logit features from the penultimate layer. This allows for exploiting an enormous amount of  engineering effort ``for free'' (e.g., network  design).

%\subsection{For Scoring: Discriminator vs. Reconstruction}
{\bf Classification vs. Reconstruction}.
We note that most, if not all, GAN-based methods largely rely on the reconstruction error for open-set discrimination~\cite{schlegl2017unsupervised, sabokrou2018adversarially, pidhorskyi2018generative, zenati2018adversarially, OzaP19}.
The underlying assumption is that closed-set data  produces lower reconstruction error than the open-set.
While this seems reasonable, it is challenging to reconstruct complex, high-resolution images~\cite{arjovsky2017wasserstein, brock2018large},
like the Cityscapes images shown in Fig.~\ref{fig:openset-pixel}.
On the contrary, %for open-set discrimination, 
we directly use the discriminator as the open-set likelihood function. 
Crucially, such a baseline has been shown to perform poorly in large body of prior work~\cite{schlegl2017unsupervised, sabokrou2018adversarially, pidhorskyi2018generative, zenati2018adversarially, liu2019generative}.
To the best of our knowledge, ours is the first result to demonstrate the strong performance of GAN-discriminators, thanks in large part to model selection via open validation (Section~\ref{ssec:opengan_method}).

\section{Experiment}
We conduct extensive experiments to validate OpenGAN under various setups, and justify the advantage of exploiting OTS features and using the GAN-discriminator as the open-set likelihood function.
We first briefly introduce three experimental setups below (details in later sections).

\begin{itemize}[noitemsep, topsep=-2pt, leftmargin=*]
\itemsep0pt
\item {\em Setup-I} open-set discrimination splits a {\em single dataset} into open and closed sets w.r.t class labels, e.g., define MNIST digits 0-5 as the closed-set for training, and digits 6-9 as the open-set in testing. Although small-scale, this is a common experimental protocol for open-set discrimination that classifies open-vs-closed test examples~\cite{neal2018open, OzaP19, perera2019deep, zhang2020hybrid}.
\item {\em  Setup-II} open-set recognition requires both $K$-way classification on the closed-set and open-set discrimination. 
%identification of the ($K$+1) open-set. 
We follow a ``less biased'' protocol~\cite{ShafaeiSL19bmvc} that constructs the open train\&test-sets with {\em cross-dataset} images~\cite{torralba2011unbiased}.
\item {\em Setup-III} examines
    the open-set discrimination at pixel level in semantic segmentation, which evaluates pixel-level open-vs-closed classification accuracy~\cite{blum2019fishyscapes, hendrycks2019benchmark}. 
    \vspace{.5mm}
\end{itemize}

{
\setlength{\tabcolsep}{0.15em} % for the horizontal padding
\begin{table*}[t]
%\scriptsize
\footnotesize
%\small
\centering
\caption[Caption for LOF]{\small
{\bf Open-set discrimination (Setup-I)} measured by  area under ROC curve (AUROC)$\uparrow$. We report methods marked by $^*$ with their best reported numbers in the compared papers.
Recall that OpenGAN-0 does not train on outlier data (i.e., $\lambda_0$=0 in Eq.~\ref{eq:OpenGAN}) and only selects discriminator checkpoints on the validation set. OpenGAN-0$^{fea}$ clearly performs the best. Defined on the off-the-shelf (OTS) features of closed-world $K$-way networks, NN$^{fea}$ and OpenGAN-0$^{fea}$ work much better than their pixel version (NN$^{pix}$ and OpenGAN-0$^{pix}$).
}
\vspace{-3mm}
\begin{tabular}{l | c c c c c c c c c c c c c c c c c c}
%\toprule
\hline
& \textcolor{black}{MSP} 
% & \textcolor{black}{Entropy} 
& {MSP$_{c}$}& \textcolor{black}{MCdrop} & \textcolor{black}{GDM}
& \textcolor{black}{OpenMax}
& \textcolor{black}{{GOpenMax}} & \textcolor{black}{{OSRCI}}
%& \textcolor{brown}{BiGAN${^{pix}_r}$} & \textcolor{brown}{BiGAN${^{pix}_d}$}
% &  \textcolor{blue}{GMM}
& \textcolor{black}{C2AE}
& CROSR
% & CGDL 
& RPL 
& Hybrid
& GDFR
& \textcolor{black}{NN$^{pix}$}
& \textcolor{black}{NN$^{fea}$}  
& \textcolor{black}{OpenGAN}
& \textcolor{blue}{OpenGAN} \\
Dataset &  \cite{hendrycks2016baseline}  
% & \cite{steinhardt2016unsupervised}  
& \cite{liang2017enhancing}
& \cite{gal2016dropout}
& \cite{lee2018simple}
& \cite{bendale2016towards}
& \ \  \cite{ge2017generative}$^*$  
& \ \  \cite{neal2018open}$^*$
% & \cite{kong2021an}  
& \ \ \cite{OzaP19}$^*$
& \ \ \cite{Yoshihashi19}$^*$ %CROSR
% & \ \ \cite{sun2020conditional}%CGDL
& \ \ \cite{chen2020learning}$^*$ 
& \ \ \cite{zhang2020hybrid}$^*$
& \ \ \cite{perera2020generative}$^*$
& \cite{ramaswamy2000efficient}
&  \cite{ramaswamy2000efficient}
& -0$^{pix}$
& \textcolor{blue}{-0$^{fea}$}\\
\hline
\emph{MNIST}  & .977  
% & .988%entropy  
& .985 & .984%MCdrop
& .989%GDM
& .981%OpenMax
&.984  & .988
%& .976 & .986
% & .993%GMM
& .989%C2AE
& .991%CROSR
% & .994%CGDL
& .996
& .995
& ---%GDFR
& .931 
& .981 & .987 & \textbf{.999}\\
\hline
\emph{SVHN}   &  .886 
% & .895%entropy 
& .891 
& .884%MCdrop
& .866%GDM
& .894%OpenMax
& \textcolor{black}{.896}  & \textcolor{black}{.910}
%& .522 & .880
% & .914%GMM 
& .922%C2AE
& .899%CROSR
% & .935%CGDL
& .968
& .947
& .935%GDFR
& .534 
& .888 & .881& \textbf{.988}\\
\hline
\emph{CIFAR}& .757  
% & .788%entropy
& .808
& .732%MCdrop
& .752%GDM
& .811%OpenMax
& \textcolor{black}{.675}  & \textcolor{black}{.699}
%& .524 & .973
% & .817%GMM 
& .895%C2AE
& .883%CROSR
% & .903%CGDL
& .901%RPL
& .950%Hybrid
& .807%GDFR
& .544 
& .801 & .971 & \textbf{.973}\\
\hline
\emph{TinyImgNet} & .577 
% & .xxx %entropy
& .713
& .675%MCdrop
& .712%GDM
& .576%OpenMax
& .580%GOpenMax 
& .586%OSRCI
%& .524 & .973
% & .817%GMM 
& .748%C2AE
& .589%CROSR
% & .xxx?%CGDL
& .809%RPL
& .793%Hybrid
& .608%GDFR
& .528%NN
& .692 & .795 & \textbf{.907}\\
%\bottomrule
\hline
\end{tabular}
\label{tab:toy_openset}
\vspace{-4mm}
\end{table*}
}

{\bf Implementation.}
We describe how to train the closed-world $K$-way classification networks which compute OTS features used for training OpenGAN$^{fea}$ (Fig.~\ref{fig:flowchart}d) and other methods (e.g., OpenMax~\cite{bendale2016towards} and C2AE~\cite{OzaP19}).
For training $K$-way networks under {\em Setup-I} and {\em II}, we train a ResNet18 model~\cite{he2016deep} exclusively on the closed-train-set (with $K$-way cross-entropy loss). Under {\em Setup-III}, we use HRNet~\cite{WangSCJDZLMTWLX19} as an OTS network, which is a top ranked model for semantic segmentation on Cityscapes~\cite{cordts2016cityscapes}.
We choose the penultimate/pre-logit layer of each $K$-way network to extract OTS features. Other layers also apply but we do not explore them in this work.
Over the features, we train OpenGAN$^{fea}$ discriminator (2MB), as well as the generator (2MB), with a multi-layer perceptron architecture. 
For comparison,  we also train a ground-up OpenGAN$^{pix}$ over pixels with a CNN architecture ($\sim$14MB)~\cite{zhu2017unpaired}. We train our OpenGAN models using GAN techniques~\cite{radford2015unsupervised}. Compared to the segmentation network HRNet (250MB),  OpenGAN$^{fea}$ is quite lightweight that induces minimal compute overhead. We conduct experiments with PyTorch~\cite{paszke2017automatic} on a single Titan X GPU.
Code is available at \url{https://github.com/aimerykong/OpenGAN}

{\bf Evaluation Metric}. 
To evaluate open-set discrimination that measures the open-vs-closed binary classification performance, we follow the literature~\cite{lee2018simple, OzaP19} and use the area under ROC curve (AUROC)~\cite{davis2006relationship}.
% receiver operating characteristic 
AUROC is a calibration-free and threshold-less metric, simplifying comparisons between methods and reliable in large open-closed imbalance situation.
For open-set recognition that measures ($K$+1)-way classification accuracy ($K$ closed-set classes plus the ($K$+1)$^{th}$ open-set class), we report the macro average F1-score over all the ($K$+1) classes on the valsets~\cite{scheirer2012toward, bendale2016towards}.

\subsection{Compared Methods}
We compare the following representative baselines and state-of-the-art methods for open-set recognition.

{\bf Baselines}.
    First, we explore classic generative models learned on the closed-train-set,
    including Nearest Neighbors (NNs)~\cite{ramaswamy2000efficient} and Gaussian Mixture Models (GMMs)
    which were found to perform quite well over L2-normalized OTS features~\cite{kong2021an}.
    We refer the reader to the supplemental for details of GMMs as they are strong yet underexplored baseline in the literature.
    Both models can be used for open-set discrimination by thresholding NN distances or likelihoods.
    We further examine the idea of outlier exposure~\cite{hendrycks2018deep} that learns an open-vs-closed binary classifier (CLS).
    Lastly, following classic work in semantic segmentation~\cite{everingham2015pascal},
    we evaluate a ($K$+1)-way classifier trained with outlier exposure, in which we use the softmax score corresponding to the ($K$+1)$^{th}$ ``other'' class as the open-set likelihood.

{\bf Likelihoods.}
    Many methods compute open-set likelihood on OTS features, including Max Softmax Probability (MSP)~\cite{hendrycks2016baseline} and  Entropy~\cite{steinhardt2016unsupervised} (derived from softmax probabilities), and calibrated MSP (MSP$_{c}$)~\cite{liang2017enhancing}.
    OpenMax~\cite{bendale2016towards} fits logits to Weibull distributions~\cite{scheirer2011meta}
    that recalibrate softmax outputs for open-set recognition.
    C2AE~\cite{OzaP19} learns an additional $K$-way classifier over the OTS features using reconstruction errors, which are then used as the  open-set likelihoods.
    GDM~\cite{lee2018simple} learns a Gaussian Discriminant Model on OTS features and computes open-set likelihood based on Mahalanobis distance.

{\bf Bayesian Networks.}
    Bayesian neural networks estimate uncertainties via Monte Carlo estimates (MCdrop)~\cite{gal2016dropout, loquercio2020general}. The estimated uncertainties are used as open-set likelihoods. We implement MCdrop via 500 samples.

{\bf GANs.}
    GOpenMax~\cite{ge2017generative} and OSRCI~\cite{neal2018open} train GANs to generate fake images to augment closed-set data for open-set recognition.
    Other types of GANs can also be used for open-set recognition, such as BiGANs~\cite{zenati2018adversarially}, on which we show our technical insights (e.g., training on OTS features and directly using the discriminator) also apply (Table~\ref{tab:BiGAN_open-set}).

When possible, we train the methods using their open-source code. 
We implement NN, CLS and OpenGAN on both RGB images (marked with $^{pix}$) and OTS features  (marked with $^{fea}$) for comparison.
For fair comparison, we tune all the models for all methods on the same val-sets.

\subsection{Setup-I: Open-Set Discrimination}

{\bf Datasets.} MNIST/CIFAR/SVHN/TinyImageNet are widely used in the open-set literature, and we follow the literature to experiment with these datasets~\cite{neal2018open, OzaP19}.
For each of the first three datasets that have ten classes, we randomly split 6 (4) classes of train/val-sets as the closed (open) train/val-sets respectively. For TinyImageNet that has 200 classes, we randomly split 20 (180) classes of train/val-sets as the closed (open) train/val-set.
On each dataset and for each method, we repeat five times with different random splits and report the average AUROC on the val-set~\cite{neal2018open, OzaP19}. As all methods have small standard deviations in their performance ($<$0.02), we omit them for brevity.

{\bf Results}.
As this setup assumes no open training data,
we cannot train discriminative classifiers like CLS. But we can still train OpenGAN-0 that uses GAN-discriminator (with model selection) as the open likelihood function. 
We have two salient conclusions from the results in Table~\ref{tab:toy_openset}. (1) Methods (e.g., NN and OpenGAN) work better on OTS features than pixels, suggesting that OTS features computed by the underlying $K$-way network are already good representations for open-set recognition.
(2) OpenGAN-0$^{fea}$ performs the best and OpenGAN-0$^{pix}$  is competitive with prior methods such as GDM and GMM, suggesting that the GAN-discriminator is a powerful open likelihood function.
% once carefully selected.

{
\setlength\tabcolsep{0.65em}
\begin{table}
\centering
\captionof{table}{\small
Our technical insights apply to other GAN-based open-set discrimination methods:
1) using BiGAN-discriminator as the open likelihood function works better than using reconstruction errors (BiGAN$_d^{fea}$ vs. BiGAN$_r^{fea}$), and 2) learning BiGANs on OTS features works much better than pixels (BiGAN$_d^{fea}$ vs.  BiGAN$_d^{pix}$).
The results are comparable to Table~\ref{tab:toy_openset}.
}
% \small
%\scriptsize
\footnotesize
\vspace{-3mm}
\begin{tabular}{l| c c | c c }
\hline
dataset & BiGAN$_r^{pix}$ & BiGAN$_r^{fea}$ & BiGAN$_d^{pix}$ & BiGAN$_d^{fea}$ \\%& GAN$^{pix}$ & GAN$^{fea}$ \\
\hline
MNIST   & .976 & .998 & .986 & .999 \\%& .987  & .999 \\
\hline
SVHN    & .822 & .976 & .880 & .993 \\%& .981  & .988 \\
\hline
CIFAR   & .924 & .967 & .968 & .973 \\%& .971  & .973 \\
\hline
\end{tabular}
\label{tab:BiGAN_open-set}
\vspace{-4.5mm}
\end{table}
}

{
\setlength{\tabcolsep}{0.23em} % for the horizontal padding
\begin{table*}[t]
% \small
%\scriptsize
\footnotesize
%\small
%\tiny
\centering
\caption[Caption for LOF]{\small
{\bf Open-set recognition (Setup-II)} measured by  
{\setlength{\fboxsep}{0pt}\colorbox{pink!35}{AUROC}}$\uparrow$, and
{\setlength{\fboxsep}{0pt}\colorbox{blue!15}{macro-averaged F1-score}}$\uparrow$ over all ($K$+1) classes.  We use TinyImageNet ($K$=200) as the closed-set, and four different datasets as the open-sets. To report a method on a specific open-test-set out of four (first column), we perform four runs in which we use one of the four datasets as a validation set for training/tuning, and then average the performance measures over the four runs with a {\setlength{\fboxsep}{0pt}\colorbox{green!20}{superscript}} marking the standard deviation. Methods such as Nearest Neighbor (NN) do not need tuning and hence have zero deviations.
We provide a summary number in the bottom macro row by averaging the results over all open-test-sets. 
Detailed results in Table~\ref{tab:generalizability}.
Clearly, a binary classifier trained on features (CLS$^{fea}$) already outperforms prior methods. However, when trained on pixels, CLS$^{pix}$ works poorly in AUROC due to overfitting to high-dimensional raw images, but performs decently in F1. To note, without handling the open-set, the $K$-way model (trained only on the closed-set TinyImageNet) achieves 0.553 F1-score over ($K$+1) classes, suggesting that, when $K$ is large ($K$=200 here), F1-score can hardly reflect open-set discrimination performance which is better measured by AUROC. 
While largely underexplored in the literature, training a ($K$+1)-way model works quite well. Clearly, OpenGAN$^{fea}$ works the best in both AUROC and F1-score.
Please refer to Fig.~\ref{fig:visualization-heterogeneous-open-image}(f-i) for ROC curves, and F1-scores vs. thresholds on the open-set likelihood.
}
\vspace{-3mm}
\begin{tabular}{l | l | c c c c c c c c  | c c c l l c c c}
%\toprule
\hline
% & OTS 
& & MSP 
% & {Entropy} 
& {OpenMax}
& {NN$^{fea}$}
& GMM
& \textcolor{black}{C2AE}
%& \textcolor{darkblue}{BiGAN$^{pix}_d$}
%& \textcolor{darkblue}{CLS$^{pix}$}
& {MSP$_{c}$} & MCdrop & GDM
& \textcolor{darkblue}{CLS$^{pix}$} & \textcolor{darkblue}{($K$+1)}
& \textcolor{darkblue}{CLS$^{fea}$}
%& \textcolor{darkblue}{GAN$^{pix}$} & \textcolor{darkblue}{GAN$^{fea}$}
& \textcolor{darkblue}{Open} & \textcolor{darkblue}{Open}
\\
\emph{open-test} 
& \emph{metric}
% & OTS 
& \cite{hendrycks2016baseline} 
%& \cite{steinhardt2016unsupervised} 
& \cite{bendale2016towards}
& \cite{ramaswamy2000efficient}
& \cite{kong2021an}
& \cite{OzaP19}
& \cite{liang2017enhancing} &  \cite{gal2016dropout}  & \cite{lee2018simple}
& \textcolor{darkblue}{\emph{}} & \textcolor{darkblue}{}
&  \textcolor{darkblue}{\emph{}} 
& \textcolor{darkblue}{GAN$^{pix}$}
% &  & % this row is GAN
&  \textcolor{darkblue}{GAN$^{fea}$}
\\
\hline
\multirow{2}{*}{\emph{CIFAR}}
& \cellcolor{pink!35}{\scriptsize  AUROC}
% & n/a
&\cellcolor{pink!35}.769\color{darkgreen}{$^{.000}$}  
&\cellcolor{pink!35}.669\color{darkgreen}{$^{.011}$} %OpenMax
&\cellcolor{pink!35}.927\color{darkgreen}{$^{.000}$} % NN
&\cellcolor{pink!35}.961\color{darkgreen}{$^{.013}$} %GMM
&\cellcolor{pink!35}.767\color{darkgreen}{$^{.020}$} % C2AE
&\cellcolor{pink!35}.791\color{darkgreen}{$^{.007}$}  % MSPc
&\cellcolor{pink!35}.809\color{darkgreen}{$^{.005}$}  %MCdrop
&\cellcolor{pink!35}.961\color{darkgreen}{$^{.007}$} %GDM
&\cellcolor{pink!35}.754\color{darkgreen}{$^{.367}$}  %OE-pix
&\cellcolor{pink!35}.880\color{darkgreen}{$^{.091}$} 
&\cellcolor{pink!35}.928\color{darkgreen}{$^{.113}$}  % OE-fea
&\cellcolor{pink!35}.981\color{darkgreen}{$^{.027}$}   %OpenGAN-pix
&\cellcolor{pink!35}.980\color{darkgreen}{$^{.011}$}  % OpenGAN
\\  %%%%%%%%%%%%% macro-F1 
& \cellcolor{blue!20}{\scriptsize  F1}
% &.xxx %OTS 
& \cellcolor{blue!20}.548\color{darkgreen}{$^{.002}$}  % MSP 
&\cellcolor{blue!20}.507\color{darkgreen}{$^{.001}$}  %OpenMax
&\cellcolor{blue!20}.525\color{darkgreen}{$^{.000}$} % NN
&\cellcolor{blue!20}.544\color{darkgreen}{$^{.002}$} %GMM
&\cellcolor{blue!20}.564\color{darkgreen}{$^{.002}$} % C2AE
&\cellcolor{blue!20}.553\color{darkgreen}{$^{.003}$}  % MSPc
&\cellcolor{blue!20}.564\color{darkgreen}{$^{.001}$}  % MCdrop
&\cellcolor{blue!20}.519\color{darkgreen}{$^{.003}$} %GDM
&\cellcolor{blue!20}.545\color{darkgreen}{$^{.032}$}  %OE-pix
&\cellcolor{blue!20}.558\color{darkgreen}{$^{.017}$} %K+1
&\cellcolor{blue!20}.555\color{darkgreen}{$^{.027}$} % OE-fea
&\cellcolor{blue!20}.563\color{darkgreen}{$^{.035}$}   %OpenGAN-pix
&\cellcolor{blue!20}.585\color{darkgreen}{$^{.003}$}  % OpenGAN
\\
\hline
\multirow{2}{*}{\emph{SVHN}}
& \cellcolor{pink!35}{\scriptsize  AUROC}
% & n/a %OTS 
&\cellcolor{pink!35}.695\color{darkgreen}{$^{.000}$}  
&\cellcolor{pink!35}.691\color{darkgreen}{$^{.014}$} %OpenMax
&\cellcolor{pink!35}.994\color{darkgreen}{$^{.000}$}  % NN
&\cellcolor{pink!35}.990\color{darkgreen}{$^{.016}$} %GMM
&\cellcolor{pink!35}.657\color{darkgreen}{$^{.018}$} % C2AE
&\cellcolor{pink!35}.863\color{darkgreen}{$^{.013}$}  % MSPc
&\cellcolor{pink!35}.783\color{darkgreen}{$^{.009}$} % MCdrop
&\cellcolor{pink!35}.999\color{darkgreen}{$^{.006}$} %GDM
&\cellcolor{pink!35}.701\color{darkgreen}{$^{.224}$}  %OE-pix
&\cellcolor{pink!35}.948\color{darkgreen}{$^{068}$} 
&\cellcolor{pink!35}.955\color{darkgreen}{$^{.052}$} % OE-fea
&\cellcolor{pink!35}.980\color{darkgreen}{$^{.014}$}   %OpenGAN-pix
&\cellcolor{pink!35}.991\color{darkgreen}{$^{.013}$} %
\\  %%%%%%%%%%%%% macro-F1 
& \cellcolor{blue!20}{\scriptsize  F1}
% & .xxx %OTS 
&\cellcolor{blue!20}.567\color{darkgreen}{$^{.002}$}   % MSP
&\cellcolor{blue!20}.551\color{darkgreen}{$^{.002}$}  %OpenMax
&\cellcolor{blue!20}.545\color{darkgreen}{$^{.000}$} % NN
&\cellcolor{blue!20}.574\color{darkgreen}{$^{.002}$} %GMM
&\cellcolor{blue!20}.565\color{darkgreen}{$^{.001}$} % C2AE
&\cellcolor{blue!20}.572\color{darkgreen}{$^{.002}$}  % MSPc
&\cellcolor{blue!20}.572\color{darkgreen}{$^{.001}$}  % MCdrop
&\cellcolor{blue!20}.575\color{darkgreen}{$^{.002}$} %GDM
&\cellcolor{blue!20}.572\color{darkgreen}{$^{.027}$}  %OE-pix
&\cellcolor{blue!20}.564\color{darkgreen}{$^{.015}$} %
&\cellcolor{blue!20}.578\color{darkgreen}{$^{.014}$} % OE-fea
&\cellcolor{blue!20}.574\color{darkgreen}{$^{.009}$}   %OpenGAN-pix
&\cellcolor{blue!20}.583\color{darkgreen}{$^{.008}$}  % OpenGAN
\\
\hline
\multirow{2}{*}{\emph{MNIST}}
& \cellcolor{pink!35}{\scriptsize  AUROC}
% & n/a % OTS
&\cellcolor{pink!35}.764\color{darkgreen}{$^{.000}$}  %
&\cellcolor{pink!35}.690\color{darkgreen}{$^{.019}$}  %OpenMax
&\cellcolor{pink!35}.901\color{darkgreen}{$^{.000}$} % NN
&\cellcolor{pink!35}.964\color{darkgreen}{$^{.021}$} %GMM
&\cellcolor{pink!35}.755\color{darkgreen}{$^{.008}$} % C2AE
&\cellcolor{pink!35}.832\color{darkgreen}{$^{.017}$}  % MSPc
&\cellcolor{pink!35}.801\color{darkgreen}{$^{.009}$}  % MCdrop
&\cellcolor{pink!35}.957\color{darkgreen}{$^{.007}$} %GDM
&\cellcolor{pink!35}.986\color{darkgreen}{$^{.327}$}  %OE-pix
&\cellcolor{pink!35}.944\color{darkgreen}{$^{.015}$} 
&\cellcolor{pink!35}.961\color{darkgreen}{$^{.083}$} % OE-fea
&\cellcolor{pink!35}.983\color{darkgreen}{$^{.068}$}  %OpenGAN-pix
&\cellcolor{pink!35}.989\color{darkgreen}{$^{.014}$} 
\\  %%%%%%%%%%%%% macro-F1 
& \cellcolor{blue!20}{\scriptsize  F1}
% & OTS 
&\cellcolor{blue!20}.559\color{darkgreen}{$^{.001}$}   % MSP
&\cellcolor{blue!20}.536\color{darkgreen}{$^{.013}$}  %OpenMax
&\cellcolor{blue!20}.553\color{darkgreen}{$^{.000}$} % NN
&\cellcolor{blue!20}.547\color{darkgreen}{$^{.008}$} %GMM
&\cellcolor{blue!20}.575\color{darkgreen}{$^{.001}$} % C2AE
&\cellcolor{blue!20}.564\color{darkgreen}{$^{.001}$}  % MSPc
&\cellcolor{blue!20}.563\color{darkgreen}{$^{.001}$}  % MCdrop
&\cellcolor{blue!20}.552\color{darkgreen}{$^{.002}$} %GDM
&\cellcolor{blue!20}.565\color{darkgreen}{$^{.020}$} %OE-pix 
&\cellcolor{blue!20}.586\color{darkgreen}{$^{.021}$} %
&\cellcolor{blue!20}.583\color{darkgreen}{$^{.010}$} % OE-fea
&\cellcolor{blue!20}.569\color{darkgreen}{$^{.016}$}   %OpenGAN-pix
&\cellcolor{blue!20}.582\color{darkgreen}{$^{.005}$}  % OpenGAN
\\
\hline
\multirow{2}{*}{\emph{Citysc.}}
& \cellcolor{pink!35}{\scriptsize  AUROC}
% & n/a % OTS 
&\cellcolor{pink!35}.789\color{darkgreen}{$^{.000}$} 
&\cellcolor{pink!35}.693\color{darkgreen}{$^{.021}$} 
&\cellcolor{pink!35}.715\color{darkgreen}{$^{.000}$}  % NN
&\cellcolor{pink!35}.867\color{darkgreen}{$^{.016}$} %GMM
&\cellcolor{pink!35}.814\color{darkgreen}{$^{.010}$} % C2AE
&\cellcolor{pink!35}.851\color{darkgreen}{$^{.003}$} % MSPc
&\cellcolor{pink!35}.868\color{darkgreen}{$^{.003}$} % MCdrop
&\cellcolor{pink!35}.513\color{darkgreen}{$^{.005}$} %GDM
&\cellcolor{pink!35}.646\color{darkgreen}{$^{.332}$} %OE-pix
&\cellcolor{pink!35}.971\color{darkgreen}{$^{.050}$} 
&\cellcolor{pink!35}.828\color{darkgreen}{$^{.032}$} % OE-fea
%& .999 & .946
&\cellcolor{pink!35}.933\color{darkgreen}{$^{.026}$} 
&\cellcolor{pink!35}.978\color{darkgreen}{$^{.013}$}  % OpenGAN-pix
\\  %%%%%%%%%%%%% macro-F1 
& \cellcolor{blue!20}{\scriptsize  F1}
&\cellcolor{blue!20}.579\color{darkgreen}{$^{.002}$}   % MSP
&\cellcolor{blue!20}.514\color{darkgreen}{$^{.002}$}  %OpenMax
&\cellcolor{blue!20}.583\color{darkgreen}{$^{.000}$} % NN
&\cellcolor{blue!20}.572\color{darkgreen}{$^{.003}$} %GMM
&\cellcolor{blue!20}.589\color{darkgreen}{$^{.002}$} % C2AE
&\cellcolor{blue!20}.583\color{darkgreen}{$^{.001}$}  % MSPc
&\cellcolor{blue!20}.571\color{darkgreen}{$^{.001}$}  % MCdrop
&\cellcolor{blue!20}.546\color{darkgreen}{$^{.003}$} %GDM
&\cellcolor{blue!20}.589\color{darkgreen}{$^{.007}$}   %OE-pix
&\cellcolor{blue!20}.561\color{darkgreen}{$^{.029}$}
&\cellcolor{blue!20}.587\color{darkgreen}{$^{.006}$} % OE-fea
&\cellcolor{blue!20}.588\color{darkgreen}{$^{.007}$}   %OpenGAN-pix
&\cellcolor{blue!20}.587\color{darkgreen}{$^{.000}$}  % OpenGAN
\\
\hline\hline
\multirow{2}{*}{\emph{average}}
& \cellcolor{pink!35}{\scriptsize  AUROC}
% & n/a
&\cellcolor{pink!35}.754\color{darkgreen}{\ \ \ \ \ \ \ } 
&\cellcolor{pink!35}.686\color{darkgreen}{\ \ \ \ \ \ \ } 
&\cellcolor{pink!35}.884\color{darkgreen}{\ \ \ \ \ \ \ } % NN
&\cellcolor{pink!35}.945\color{darkgreen}{\ \ \ \ \ \ \ } %GMM
&\cellcolor{pink!35}.748\color{darkgreen}{\ \ \ \ \ \ \ } % C2AE
&\cellcolor{pink!35}.834\color{darkgreen}{\ \ \ \ \ \ \ } 
&\cellcolor{pink!35}.815\color{darkgreen}{\ \ \ \ \ \ \ }  % MCdrop
&\cellcolor{pink!35}.857\color{darkgreen}{\ \ \ \ \ \ \ } %GDM
&\cellcolor{pink!35}.772\color{darkgreen}{\ \ \ \ \ \ \ } %OE-pix
&\cellcolor{pink!35}.936\color{darkgreen}{\ \ \ \ \ \ \ } 
&\cellcolor{pink!35}.918\color{darkgreen}{\ \ \ \ \ \ \ } % OE-fea
&\cellcolor{pink!35}.969\color{darkgreen}{\ \ \ \ \ \ \ }  %OpenGAN-pix
&\cellcolor{pink!35}\textbf{.984}\color{darkgreen}{\ \ \ \ \ \ \ } 
\\  %%%%%%%%%%%%% macro-F1 
& \cellcolor{blue!20}{\scriptsize  F1}
%\bottomrule
% & OTS
&\cellcolor{blue!20}.560\color{darkgreen}{\ \ \ \ \ \ \ }  
%& .784  
&\cellcolor{blue!20}.527\color{darkgreen}{\ \ \ \ \ \ \ }  %OpenMax
&\cellcolor{blue!20}.552\color{darkgreen}{\ \ \ \ \ \ \ } % NN
&\cellcolor{blue!20}.559\color{darkgreen}{\ \ \ \ \ \ \ } %GMM
&\cellcolor{blue!20}.569\color{darkgreen}{\ \ \ \ \ \ \ } % C2AE
&\cellcolor{blue!20}.568\color{darkgreen}{\ \ \ \ \ \ \ } % MSPc
&\cellcolor{blue!20}.567\color{darkgreen}{\ \ \ \ \ \ \ }  % MCdrop
&\cellcolor{blue!20}.548\color{darkgreen}{\ \ \ \ \ \ \ } %GDM
&\cellcolor{blue!20}.568\color{darkgreen}{\ \ \ \ \ \ \ } %OE-pix  
&\cellcolor{blue!20}.565\color{darkgreen}{\ \ \ \ \ \ \ } %K+1
&\cellcolor{blue!20}.576\color{darkgreen}{\ \ \ \ \ \ \ }
&\cellcolor{blue!20}.573\color{darkgreen}{\ \ \ \ \ \ \ }   %OpenGAN-pix
&\cellcolor{blue!20}\textbf{.584}\color{darkgreen}{\ \ \ \ \ \ \ }  % OpenGAN-fea
\\
\hline
\end{tabular}
\label{tab:heterogeneous-open-image}
\vspace{-1mm}
\end{table*}
}

\begin{figure*}[t]
\centering
\includegraphics[width=0.99\linewidth]{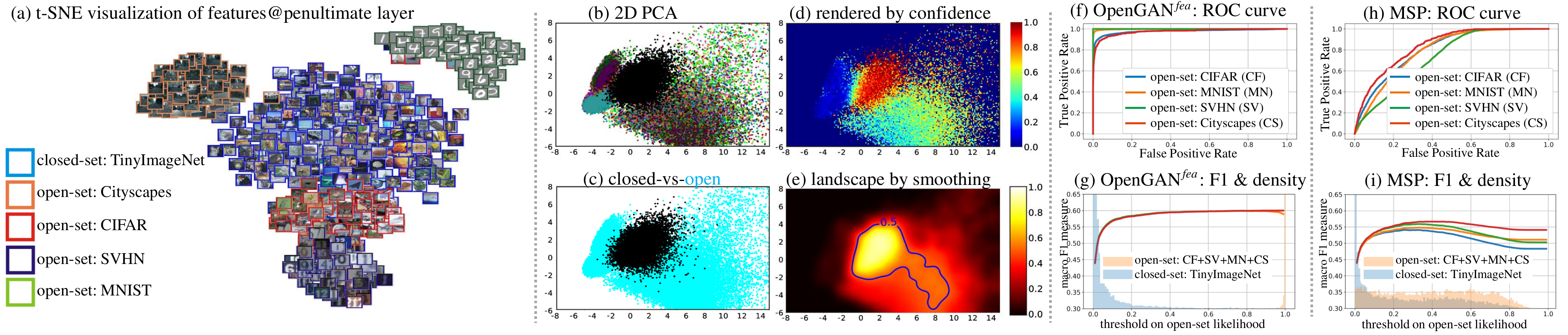}
\vspace*{-4.5mm}
\caption{\small
% {\bf Landscape of the OpenGAN classifier.}
{\bf (a)} We use t-SNE to visualize the embedding space through the OTS features computed by the $K$-way  network trained on TinyImageNet train-set. Images from the other datasets are open-set examples. Clearly, closed and open examples are well separated in the feature space.
We further visualize the ``landscape'' of the OpenGAN$^{fea}$ open-set discriminator, by {\bf (b)} projecting the OTS features into 2D using PCA; 
{\bf (c)} coloring them with their closed/open labels; 
{\bf (d)} rendering them with their open-set likelihoods computed by OpenGAN$^{fea}$; 
{\bf (e)} smoothing with Gaussian Filtering overlaid with the OpenGAN's decision boundary.
We further compare OpenGAN$^{fea}$ (tuned on SVHN) and MSP in {\bf (f-g)} for open-set discrimination by ROC curves, and in {\bf (h-i)} for open-set recognition by curves of the F1-score vs. thresholds of open likelihold. We render the density of open and closed testing data using shadows in {\bf (g)} and {\bf (i)}. 
In these plots, we use each of the four cross-dataset open-test-sets (unseen in training) as an independent open-set to draw the curves.
% While Table~\ref{tab:heterogeneous-open-image} shows MSP performs well, 
The curves clearly show that OpenGAN significantly outperforms MSP on open-set discrimination (AUROC) and open-set  recognition (F1). 
}
\vspace{-4mm}
\label{fig:visualization-heterogeneous-open-image}
\end{figure*}

{\bf Further Analysis}.
There are many other GAN-based open-set methods, such as training BiGANs~\cite{schlegl2017unsupervised, zenati2018efficient, zenati2018adversarially} or adversarial autoencoders~\cite{pidhorskyi2018generative, sabokrou2018adversarially} on raw images, and using the reconstruction error 
%(on either images or encoded features) 
as open-set likelihood~\cite{schlegl2017unsupervised, sabokrou2018adversarially,  zenati2018adversarially, akcay2018ganomaly, deecke2018image}.
We show our technical insights apply to different GAN architectures for open-set recognition:
(1) using GAN-discriminator as the open-set likelihood function instead of pixel reconstruction errors, and (2) training them on OTS features rather than raw pixels.
We hereby analyze a typical BiGAN-based method~\cite{zenati2018adversarially},
which learns a BiGAN with both the reconstruction error and the GAN-discriminator. 
We compare BiGAN's performance by either using the reconstruction error (BiGAN$_{r}$) or its discriminator (BiGAN$_{d}$) for open-set recognition.
%{\em itself} (as we advocate) as the open-set likelihood for open-set recognition, we compare the two and demonstrate our insight-1.
We also compare building BiGANs on either pixels (BiGAN$^{pix}$) or features (BiGAN$^{fea}$).
Table~\ref{tab:BiGAN_open-set} lists detailed comparisons under {\em Setup-I}  (all models are selected on the val-sets).
Clearly, our conclusions hold regardless of the base GAN architecture: 1) using OTS features rather than pixels (cf. BiGAN$^{fea}$ vs BiGAN$^{pix}$), and 2) more importantly, using discriminators instead of reconstruction errors (cf. BiGAN$_{d}$ vs. BiGAN$_{r}$).
% }

\subsection{Setup-II: Cross-Dataset Open-Set Recognition}
\label{ssec:hetergoneous_open_image_discrimination}

Using cross-dataset examples as the open-set is another established protocol~\cite{liang2017enhancing, lee2018simple, hendrycks2018deep, dhamija2018reducing}. We follow the ``less biased'' protocol introduced in \cite{ShafaeiSL19bmvc}, which uses three datasets for benchmarking that reduces dataset-level bias~\cite{torralba2011unbiased}.
This protocol tests the generalization of open-set methods to diverse open testing examples.

{\bf Datasets}. We use TinyImageNet as the closed-set for $K$-way classification ($K$=200). 
% , which has 200 classes of 64x64 images. 
Images of each class are split into 500/50/50 images as the train/val/test sets.
Following~\cite{ShafaeiSL19bmvc},
we construct open train/val and test sets using different datasets~\cite{torralba2011unbiased},
including MNIST (MN), SVHN (SV), CIFAR (CF) and Cityscapes (CS).
For example,
we use MNIST {\em train}-set to tune/train a model,
and test it on CIFAR {\em test}-set as open-test set.
This allows for analyzing how open-set methods generalize to diverse open testing examples (cf. Table~\ref{tab:generalizability}).
We use bilinear interpolation to resize all images to 64x64 to match TinyImageNet image resolution.

{
\setlength{\tabcolsep}{0.57em} % for the horizontal padding
\begin{table*}[t]
%\scriptsize
\footnotesize
%\small
%\tiny
%\footnotesize
\centering
\caption[Caption for LOF]{\small
{\bf Diagnostic analysis for cross-dataset open-set discrimination} measured by AUROC$\uparrow$.
In this setup, the TinyImageNet train/val/test sets serve as the closed train/val/test sets, and open train/test sets are the other two different datasets.
Following outlier exposure~\cite{hendrycks2016baseline}, we train/tune CLS and OpenGAN on a cross-dataset as the open train-set. Recall that we do not train OpenGAN-0 on {\em any} open examples, although we tune it on the respective cross-dataset open train-set.
CLS and OpenGAN use their last-epoch checkpoints to report performance.
For better comparison, we report the {\setlength{\fboxsep}{0pt}\colorbox{green!15}{average}} AUROC performance across all open-val-sets in the last column.
%indicating how each method generalizes to diverse open-test-data.
We color the entries that have \note{AUROC $<$0.9} with \note{red}, implying these models overfit to the open-train-set and generalize poorly on the other open-test-set.
OpenGAN$^{fea}$ clearly performs the best; while CLS (esp. CLS$^{pix}$ which operates on pixels) generalizes poorly. Perhaps surprisingly, OpenGAN-0 performs equally well although it does not train on open taining data.
}
\vspace{-3mm}
%\begin{tabular}{l | c | c | c | c | c | c | c | c | c | c | c | c | c | c | c | c }
\begin{tabular}{l   c c c c   c c c c   c c c c   c c c c   c   }
\toprule
\emph{open-val-set}
    & \multicolumn{4}{c}{CIFAR10 (CF)}
    & \multicolumn{4}{c}{SVHN (SV)}
    & \multicolumn{4}{c}{MNIST (MN)}
    &  \multicolumn{4}{c}{Cityscapes (CS)}  &  \multirow{2}{*}{\em avg.}  \\
%\hline
\cmidrule(lr){2-5} \cmidrule(lr){6-9} \cmidrule(lr){10-13}  \cmidrule(lr){14-17} % \cmidrule{9-12} \cmidrule{15-18}  \cmidrule{21-24}
\emph{open-test-set}
    &  CF & SV& MN& CS
    &  CF & SV& MN& CS
    &  CF & SV& MN& CS
    &  CF & SV& MN& CS & \\
\midrule
CLS$^{pix}$ &  .999&.999 &\note{.101}& .895
    &  .935 & .999 & \note{.453} & .972
    &  \note{.411} & \note{.340} & .999 & \note{.113}
    &  \note{.317} & \note{.512} & \note{.100} & .999 & \cellcolor{green!15}.634 \\
OpenGAN-0$^{pix}$ & .999 & .998 & \note{.550} & .999
    & .999 & .999 & .993 & .999
    & .999 & .968 & .999 & .911
    & .999 & .999 & .915 & .999 & \cellcolor{green!15}.958\\
OpenGAN$^{pix}$  & .999 & .999 & .989 & .933
    & .974 & .999 & .997 & .967
    & .976 & .998 & .999 & \note{.835}
    & .967 & .928 & .950 & .999 & \cellcolor{green!15}.969 \\
\hline
CLS$^{fea}$ & .999 & .933 & .916 & \note{.699}
    & .940 & .999 & .979 & \note{.863}
    & \note{.893} & .961 & .999 & \note{.781}
    & \note{.881} & .926 & .949 & .968 & \cellcolor{green!15}.918\\
OpenGAN-0$^{fea}$ & .999 & .998 & .997 & .999
    & .964 & .996 & .996 & .946
    & .952 & .992 & .994 & .934
    & .994 & .995 & .992 & .997 & \cellcolor{green!15}{\bf .984}\\
OpenGAN$^{fea}$  & .999 & .999 & .990 & .973
    & .974 & .999 & .996 & .971
    & .976 & .998 & .999 & .967
    & .973 & .968 & .970 & .999 & \cellcolor{green!15}\textbf{.984} \\
%\hline
\bottomrule
\end{tabular}
\label{tab:generalizability}
\vspace{-5mm}
\end{table*}
}

{\bf Results}.
Table~\ref{tab:heterogeneous-open-image} shows detailed results.
First, methods perform much better on features than pixels
(e.g., CLS$^{fea}$ vs. CLS$^{pix}$);
and our OpenGAN performs the best.
Perhaps surprisingly,
OpenMax, a classic open-set,
does not work well in this setup.
This is consistent with the results in~\cite{dhamija2018reducing, ShafaeiSL19bmvc}.
We conjecture that OpenMax cannot effectively recognize cross-dataset open-set examples represented by logit features (computed by the $K$-way network) which are too invariant to be adequate for open-set recognition.
Moreover, the ($K$+1)-way classifier also works quite well,
even outperforming the open-vs-closed binary classifiers (CLS) in AUROC.
Next we analyze why the  binary classifier CLS (as widely done since~\cite{hendrycks2018deep}) are less effective.

{\bf Further Analysis}.
Table~\ref{tab:generalizability} lists detailed results of OpenGAN, CLS ($\lambda_G$=0 in Eq.~\ref{eq:OpenGAN}) and OpenGAN-0 ($\lambda_o$=0 in Eq.~\ref{eq:OpenGAN}), when trained/tuned and tested on different cross-dataset open-set examples.
All methods perform better on OTS features than pixels (cf. CLS$^{fea}$ vs. CLS$^{pix}$);
and work almost perfectly when trained and tested with the same open-set dataset,
e.g., column-cf under ``CIFAR-train (cf)'' where we use CIFAR images as the open-set data.
However, when tested on a different dataset of open-set examples,
CLS performs quite poorly (especially when built on pixels)  because it overfits easily to  high-dimensional pixel images~\cite{ShafaeiSL19bmvc}.
In contrast, with \emph{fake} open-data generated adversarially, OpenGAN and its special form OpenGAN-0 perform and generalize much better.
Nevertheless, this implies a failure mode of OpenGAN, because the open-set data used in training could be quite different from those in testing, potentially leading to an OpenGAN that  perform poorly in the real open world.
Perhaps surprisingly, OpenGAN-0$^{fea}$ performs as well as OpenGAN$^{fea}$, although it does not train on open-set data. This further shows the merit of generating \emph{fake} open examples to augment heavily-biased open-set training data, and our technique insights (as previously analyzed under {\em Setup-I}): 1) using GAN-discriminator as the likelihood function, and 2) training GANs on OTS features rather than pixels.

{\bf Visualization}.
Fig.~\ref{fig:flowchart} shows some synthesized images by GAN$^{pix}$, and we visualize more in the supplement.
To intuitively illustrate how the synthesized images help better span the open-world,
we analyze why a simple discriminator works so well when trained on the OTS features.
We visualize the features in Fig.~\ref{fig:visualization-heterogeneous-open-image} (a) and ``decision landscape'' in Fig.~\ref{fig:visualization-heterogeneous-open-image} (b-e), demonstrating that the closed- and open-set images are clearly separated in the feature space.

\subsection{Setup-III: Open-Set Semantic Segmentation}

Open-set semantic segmentation has been explored in recent work~\cite{blum2019fishyscapes, hendrycks2019benchmark}, which creates synthetic open-set pixels by pasting virtual objects (e.g., cropped from PASCAL VOC masks~\cite{everingham2015pascal}) on Cityscapes images. In this work, we do not generate synthetic pixels but instead repurpose ``other'' pixels (outside the set of $K$ classes) that already exist in Cityscapes.
Interestingly, classic semantic segmentation benchmarks evaluate  these ``other'' pixels as a separate background class~\cite{everingham2015pascal}, but Cityscapes ignores them in its evaluation (as do many other contemporary datasets~\cite{caesar2020nuscenes, behley2019semantickitti, richter2016playing, neuhold2017mapillary}).
The historically-ignored pixels include vulnerable objects (e.g., strollers in Fig.~\ref{fig:openset-pixel}), and can be naturally evaluated as open-set examples.

{\bf Datasets}.
Cityscapes~\cite{cordts2016cityscapes} contains 1024x2048 high-resolution urban scene images with 19 class labels for semantic segmentation.
We construct our train- and val-sets from its 2,975 training images,
in which we use the last 10 images as val-set and the rest as train-set.
We use its official 500 validation images as our test-set.
The ``other'' pixels (cf. Fig.~\ref{fig:openset-pixel})
are the open-set examples in this setup.
We refer readers to the supplemental for details,
such as model architecture, batch construction, weight tuning, etc.

{
\setlength{\tabcolsep}{0.12em} % for the horizontal padding
\begin{table*}[t]
\centering
% \scriptsize
\footnotesize
\caption{\small
{\bf Comparison in open-set semantic segmentation} on Cityscapes (AUROC $\uparrow$).
All methods are implemented on top of the segmentation network HRNet~\cite{WangSCJDZLMTWLX19} except the ones operating on pixels (as marked by $^{pix}$).
Our approach OpenGAN$^{fea}$ clearly performs the best.
Fig.~\ref{fig:curves_segm} analyzes OpenGAN trained with varied number of open-set pixels, when built on either pixels or OTS features.
}
\vspace{-3mm}
\begin{tabular}{c c c c c c c c c c c c c c }
\hline
MSP \cite{hendrycks2016baseline} & Entropy \cite{steinhardt2016unsupervised} & OpenMax \cite{bendale2016towards} & C2AE \cite{OzaP19} & {MSP$_{c}$} \cite{liang2017enhancing} & {MCdrop} \cite{gal2016dropout} &  GDM \cite{lee2018simple} & GMM  \cite{kong2021an} & {HRNet-($K$+1)} & {OpenGAN-0$^{fea}$}   &  CLS$^{fea}$ &  {OpenGAN$^{fea}$}   \\  
\hline
.721 & .697 & .751 & .722 & .755 & .767 & .743 &  .765 & .755 & .709 &  .861 & {\bf .885}   \\ %& {\bf .885} \\
% \midrule
\hline
\end{tabular}
\vspace{-2mm}
\label{tab:cityscapes}
\end{table*}
}

\begin{figure*}[t]
\centering
\includegraphics[width=1\linewidth]{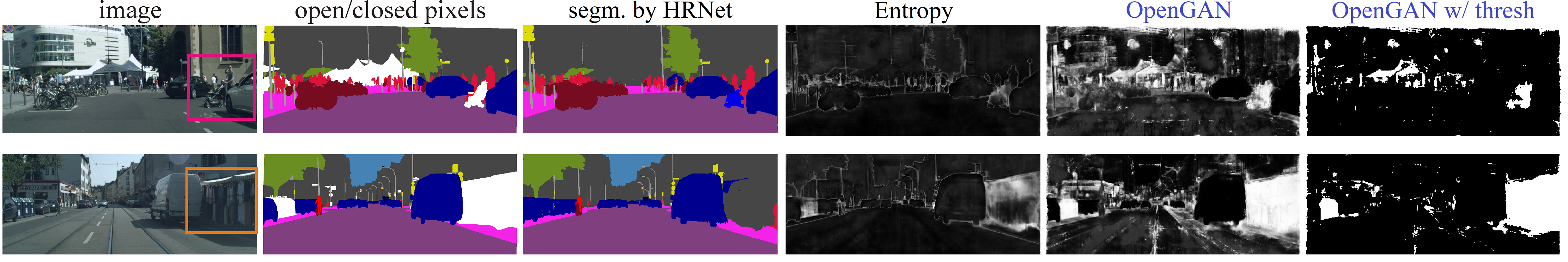}
\vspace*{-8mm}
\caption{\small
Qualitative results of two testing images, on which a state-of-the-art network (HRNet) misclassifies the \emph{unknown} categories \textcolor{red}{stroller}/\textcolor{orange}{street-shop} as motorcycle/building.
From left to right of each row:
the input image, its per-pixel semantic labels (in which {\tt white} regions are open-set pixels),  the semantic segmentation result by HRNet,
open-set likelihoods by Entropy, our OpenGAN$^{fea}$,
and its thresholded open-pixel map (threshold=0.7).
OpenGAN clearly captures most open-set pixels (the white ones). 
Note that the \textcolor{orange}{street-shop} is a real open-set example because
Cityscapes train-set does not have another \textcolor{orange}{street-shop} like this size and content (i.e., selling clothes). 
}
\vspace{-2.5mm}
\label{fig:qualitativeDemo}
\end{figure*}

\begin{figure}[t]
\centering
\includegraphics[width=1\linewidth]{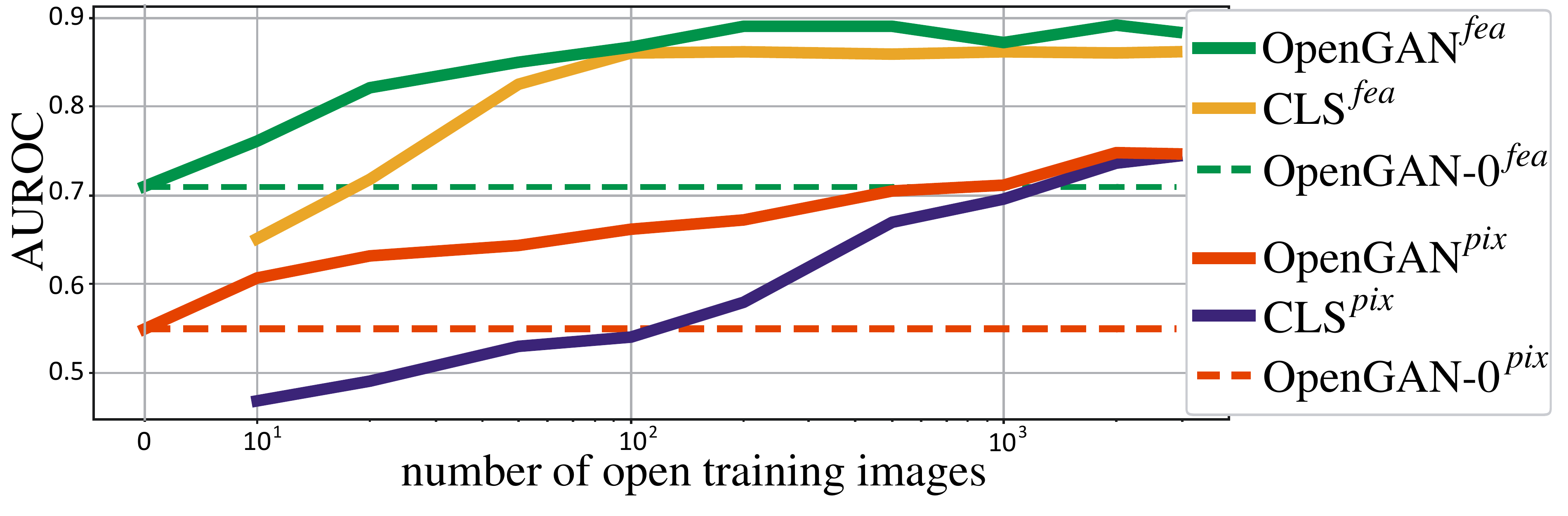}
\vspace{-8mm}
\caption{\small
Diagnostic study w.r.t AUROC vs. number of open images which provide open-set training pixels.
Our methods perform better on OTS features than pixels.
Recall OpenGAN-0 is equivalent to training a normal GAN (without open training data) and using its discriminator as open-set likelihood.
With some open training data (e.g., 100 open images), CLS outperforms OpenGAN-0; but OpenGAN consistently performs the best. 
}
\vspace{-4mm}
\label{fig:curves_segm}
\end{figure}

%A more realistic setup for open-set recognition is open-pixel classification via semantic segmentation.
{\bf Pixel Generation}. 
As Cityscapes has high-resolution images (1024x2048), it is nontrivial to train OpenGAN$^{pix}$, especially its special form OpenGAN-0$^{pix}$, which must learn to generate high-resolution images. 
We find the successful training of OpenGAN-0$^{pix}$ depends on the resolution of images to be generated: we train OpenGAN-0$^{pix}$ by generating patches (64x64), not full-resolution images.

{\bf Results}. 
Table~\ref{tab:cityscapes} lists quantitative results.
As we train OpenGAN and CLS with open pixels,
we diagnose in Fig.~\ref{fig:curves_segm} the open-set performance by varying the number of training images 
that provide the open-training pixels, along with closed-training pixels from all training images.
First, these results show that OpenGAN$^{fea}$ substantially outperforms all other methods. Generally speaking, the methods that process features outperform those that process pixels (e.g.,  OpenGAN and CLS in Fig.~\ref{fig:curves_segm}). This suggests that OTS features (from the segmentation network) serve as a powerful representation for open-set pixel recognition.
The curves in Fig.~\ref{fig:curves_segm} imply that methods with enough data on pixels should work (e.g., achieving similar performance as on features). This is consistent with evidence from semantic segmentation works.
However, methods saturate more quickly on OTS features than pixels, suggesting the benefit of using OTS features for open-set recognition.
Moreover, OpenGAN-0 performs better than CLS when trained on fewer open-training images (e.g., 10). But with modest number of open training images (e.g., 50), CLS outperforms OpenGAN-0 and other classic methods (e.g., OpenMax and C2AE in Table~\ref{tab:cityscapes}) which assume no open training data.
This confirms the effectiveness of Outlier Exposure, even with a modest amount of outliers~\cite{hendrycks2018deep}.

Bayesian networks (MCdrop and MSP$_{c}$) outperform the baseline MSP, showing that uncertainties can be reasonably used for open-set recognition.
Lastly, we train a ``ground-up'' ($K$+1)-way HRNet model that treats ``other'' pixels as the ($K$+1)$^{th}$ background class~\cite{everingham2015pascal}, shown by HRNet-($K$+1) in Table~\ref{tab:cityscapes}.
It performs better than other typical open-set methods but much lower than the simple open-vs-closed binary classifier CLS$^{fea}$, presumably because the ($K$+1)-way model has to strike a balance over all the ($K$+1) classes while the binary CLS benefits from training on more balanced batches of closed/open pixels.

{\bf Visualization}.
Fig.~\ref{fig:qualitativeDemo} qualitatively compares OpenGAN and the entropy method (more visual results are in the supplemental).
The visualization shows OpenGAN sufficiently recognize open-set pixels. It also implies failure happens when OpenGAN  misclassifies open-vs-closed pixels.
Fig.~\ref{fig:genImages_cityscapes} compares some generated patches by OpenGAN-0$^{fea}$ and OpenGAN-0$^{fea}$, intuitively showing why using OTS features leads to better performance for open-set recognition.

\begin{figure}[t]
\centering  %\fbox{\rule{0pt}{2in} \rule{0.9\linewidth}{0pt}}
\vspace{-2mm} 
\hspace{7mm} {\small Real} \hspace{15mm}  {\small OpenGAN-0$^{pix}$} \hspace{8mm}   {\small OpenGAN-0$^{fea}$}  \\
\includegraphics[width=0.98\linewidth]{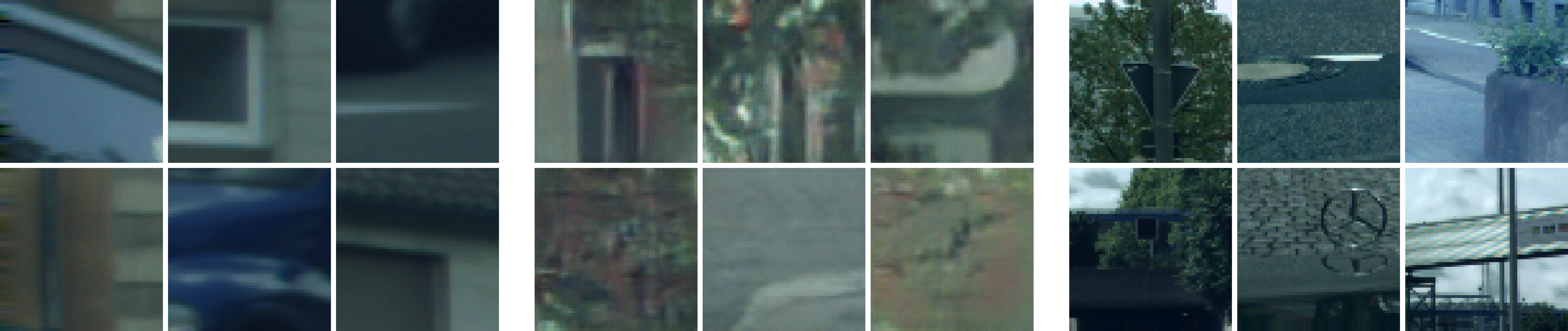}
\vspace{-3mm}
\caption{\small
    \solved{
    Visuals of Cityscapes real image patches (left), synthesized patches by OpenGAN-0$^{pix}$ (mid) and OpenGAN-0$^{fea}$ (right).
    As OpenGAN-0$^{fea}$ generates features instead of pixel patches, we ``synthesize'' the patch for a generated feature  by finding the closest pixel feature on the training set and returning its surrounding image patch.  We can see OpenGAN-0$^{pix}$ synthesizes realistic patches w.r.t color and tone, but it (0.549 AUROC) notably underperforms OpenGAN-0$^{fea}$ (0.709 AUROC) for open-set segmentation.
    The ``synthesized'' patches by OpenGAN-0$^{fea}$ capture many open-set objects, such as bridge,  back-of-traffic-sign and unknown-static-objects, none of which belong to any of the 19 closed-set classes in the Cityscapes benchmark. This intuitively shows why methods work better on OTS features than pixels.
    }
}
\vspace{-4mm}
\label{fig:genImages_cityscapes}
\end{figure}

\section{Conclusion}
We propose {\bf OpenGAN} for open-set recognition by incorporating two technical insights, 1) training an open-vs-closed classifier on OTS features rather than pixels, and 2) adversarialy synthesizing \emph{fake} open data to augment the set of open-training data. With OpenGAN, we show using GAN-discriminator \emph{does} achieve the state-of-the-art on open-set discrimination, once being selected using a val-set of real outlier examples. This is effective  even when the outlier validation examples are sparsely sampled or strongly biased.
OpenGAN significantly outperforms prior art on both open-set image recognition and semantic segmentation.

\section*{Acknowledgement}
This work was supported by the CMU Argo AI Center for Autonomous Vehicle Research.

{\small
\bibliographystyle{ieee_fullname}
\bibliography{egbib}
}

\newpage
\clearpage
\newpage
\clearpage

\section*{}
\begin{center}
{\bf \large Outline}
\end{center}
\emph{
As elaborated in the main paper,
our proposed OpenGAN trains an open-vs-closed  binary classifier for open-set recognition. Our three major technical insights are
(1) model selection of a GAN-discriminator as the open-set likelihood function via validation,
(2) augment the available set of real open training examples with adversarially synthesized ``fake” data,
and (3) training OpenGAN on off-the-shelf (OTS) features rather than pixel images.
We expand on the techniques of OpenGAN in the appendix, including architecture design,  model selection and additional details for training.
We also provide additional comparisons to recently published methods and qualitative results. Below is the outline.
}

{\bf Section~\ref{sec:model-arch}:  Model architectures}
for both OpenGAN$^{fea}$ and OpenGAN$^{pix}$.

{\bf Section~\ref{sec:setup_open-pix_experiment}: Detailed setup for open-set semantic segmentation}, such as data statistics (e.g., the number of open-set pixels in the testing set) and batch construction during training.

{\bf Section~\ref{sec:model-selection}: Model selection}
that is performed on a validation set.

{\bf Section~\ref{sec:param-tuning}: Hyper-parameter tuning} which is performed on a validation set.

{\bf Section~\ref{sec:GMM}: Statistical methods for open-set recognition} that learn generative models (e.g., Gaussian Mixture) over off-the-shelf deep features.

{\bf Section~\ref{sec:sota-comparison}: More quantitative comparisons} to several approaches published recently.
% We compare with more approaches published recently.

{\bf Section~\ref{sec:visualization}: Visuals of synthesized images} generated by OpenGAN-0$^{pix}$ and OpenGAN-0$^{fea}$, intuitively demonstrating their effectiveness and limitations.

{\bf Section~\ref{sec:open-pixel-vis}: Visual results of open-set semantic segmentation}.

{\bf Section~\ref{sec:limitation}: Failure Cases and Limitations}.

\section{Model Architecture}
\label{sec:model-arch}
We describe the network architectures of OpenGAN.
Because our final version OpenGAN$^{fea}$ operates on off-the-shelf (OTS) features, we use multi-layer perceptron (MLP) networks for the generator and discriminator.
Because OpenGAN$^{pix}$ operates on pixels, we make use of convolutional neural network (CNN) architectures.
We begin with the former.

\subsection{OpenGAN$^{fea}$ architecture}

OpenGAN$^{fea}$ consists of a generator and a discriminator.
OpenGAN$^{fea}$ is compact in terms of model size ($\sim$2MB), because it adopts MLP network over OTS features which are low-dimensional (e.g., 512-dim vectors) compared to pixel images. The MLP architectures are described below

\begin{itemize}
\item The MLP discriminator in OpenGAN$^{fea}$ takes a $D$-dimensional feature as the input. Its architecture has a set of fully-connected layers (fc marked with input-dimension and output-dimension), Batch Normalization layers (BN) and  LeakyReLU layers (hyper-parameter as 0.2):
{\tt
fc (D$\rightarrow$64*8), BN, LeakyReLU,
fc (64*8$\rightarrow$64*4), BN, LeakyReLU,
fc (64*4$\rightarrow$64*2), BN, LeakyReLU,
fc (64*2$\rightarrow$64*1), BN, LeakyReLU,
fc (64*1$\rightarrow$1), Sigmoid.}
\item The MLP generator synthesizes a $D$-dimensional feature given a 64-dimensional random vector:
{\tt fc (64$\rightarrow$64*8), BN, LeakyReLU,
fc (64*8$\rightarrow$64*4), BN, LeakyReLU,
fc (64*4$\rightarrow$64*2), BN, LeakyReLU,
fc (64*2$\rightarrow$64*4), BN, LeakyReLU,
fc (64*4$\rightarrow$D), Tanh.}
\end{itemize}

For open-set image classification, the image features have dimension $D=512$ from ResNet18 (the $K$-way classification networks under {\em Setup-I} and {\em II}).
For open-set segmentation, the per-pixel features have dimension $D=720$ at the penultimate layer of HRnet (a top-ranked  semantic segmentation model used in this work under {\em Setup-III}).

\subsection{OpenGAN$^{pix}$ architecture}

OpenGAN$^{pix}$'s generator and discriminator follow the CycleGAN architecture~\cite{zhu2017unpaired}.
We change the stride size in the convolution layers to adapt the networks to specific image resolution (e.g., CIFAR 32x32 and TinyImageNet 64x64).
The generator and discriminator in OpenGAN$^{pix}$ have model sizes as $\sim$14MB and $\sim$11MB, respectively.
We find it important to ensure that OpenGAN$^{pix}$ has a larger capacity than OpenGAN$^{fea}$ to generate high-dimensional RGB raw images.

% \newpage
\section{Setup for Open-Set Semantic Segmentation}
\label{sec:setup_open-pix_experiment}

\begin{figure}[t]
\centering
\begin{minipage}{1\linewidth}  % \textwidth-2
\centering
\includegraphics[width=1\linewidth, clip, trim={8.7cm 11.6cm 3.5cm 12.2cm}]{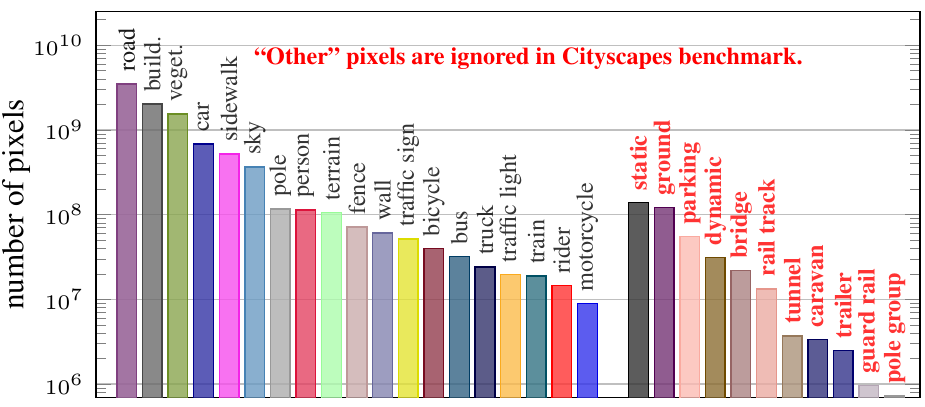}
\end{minipage}
% \end{tabular*}%
\vspace{-2.5mm}
\caption{\small
% \redo{
Cityscapes annotates a sizeable portion of pixels that do not belong to one of the $K$ closed-set classes on which the Cityscapes benchmark evaluates. As a result, many methods also ignore them during training~\cite{WangSCJDZLMTWLX19}.
% Interestingly, such pixels are not evaluated during testing, and as a result are often ignored by most methods during training.
We repurpose these historically-ignored pixels as open-set examples that are from the ($K$+1)$^{th}$ ``other'' class, allowing for a large-scale exploration of open-set recognition via semantic segmentation.
}
\vspace{-1mm}
\label{fig:histogram_in_cityscapes}
\end{figure}

\begin{figure}[t]
\centering
\includegraphics[width=\linewidth]{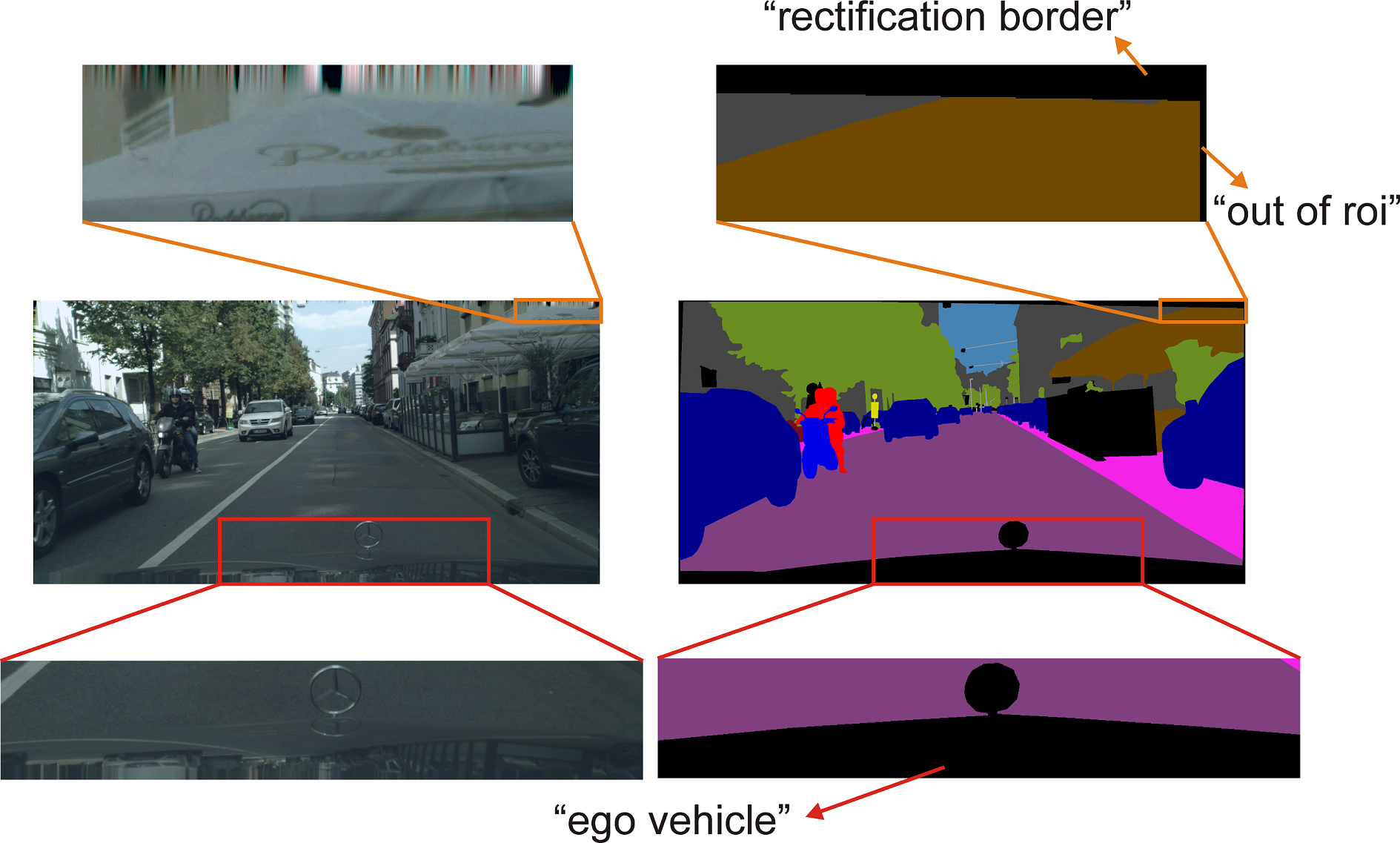}
%\vspace*{-3mm}
\caption{\small
{\bf Void pixels in Cityscapes} that are not from the closed-set classes nor open-set. We highlight these pixels over an image (left) and its semantic annotations (right). Cityscapes labels these pixels as {\tt rectification-border} (artifacts at the image borders caused by stereo rectification), {\tt ego-vehicle} (a part of the car body at the bottom of the image including car logo and hood) and {\tt out-of-roi} (narrow strip of 5 pixels along the image borders).
These noise-pixels can be easily identified without machine-learned methods.
Therefore, we do not evaluate on these pixels.
}
%\vspace{-5mm}
\label{fig:known_noise_cityscapes}
\end{figure}

In this work, we use Cityscapes to study open-set semantic segmentation. Prior work suggests pasting virtual objects (e.g., cropped from PASCAL VOC masks~\cite{everingham2015pascal}) on Cityscapes images as open-set pixels~\cite{blum2019fishyscapes, hendrycks2019benchmark}.
We notice that Cityscapes ignores a sizeable portion of pixels in its benchmark, as demonstrated by Figure~\ref{fig:histogram_in_cityscapes}. As a result, many methods also ignore them in training. Therefore, instead of introducing artificial open-set examples, we use the historically-ignored pixels in Cityscapes as the real open-set examples.
We hereby describe in detail our configuration for open-set semantic segmentation setup and experiments on Cityscapes.

{\bf Data Setup.}
Cityscapes training set has 2,975 images. We use the first  2,965 images for training, and hold out the last 10 as validation set for model selection.
We use the 500 Cityscapes validation images as our test set.
Here are the statistics for the full train/val/test sets.
\begin{itemize} % [noitemsep, topsep=-2pt] % , leftmargin=*
  \item train-set for closed-pixels: 2,965 images providing 334M closed-set pixels. % 344,615,450 (344M)
  \item train-set for open-pixels:  2,965 images providing 44M open-set pixels. % 44,321,324 (44M)
  \item val-set for closed-pixels: 10 images providing 1M  closed-set pixels. % 1,111,228 (1M)
  \item val-set for open-pixels: 10 images providing 0.2M open-set pixels. % 199,492 (0.2M)
  \item test-set for closed-pixels: 500 images providing 56M pixels. % 56,116,858 (56M)
  \item test-set for open-pixels:
  500 images providing  2M pixels.  % 2,718,339 (2M)
\end{itemize}

Note that,
we exclude the pixels labeled with {\tt rectification-border} (artifacts at the image borders caused by stereo rectification), {\tt ego-vehicle} (a part of the car body at the bottom of the image including car logo and hood) and {\tt out-of-roi} (narrow strip of 5 pixels along the image borders).  These pixels can be easily localized using camera information. We demonstrate such pixels in Figure~\ref{fig:known_noise_cityscapes}.  In this sense, these pixels are not \emph{unknown} open-set pixels but \emph{known} noises caused by sensors and viewpoint. Therefore, we do not include them for open-set evaluation.

{\bf Feature setup.}
\begin{itemize} % [noitemsep, topsep=-2pt]
  \item $M^{pix}$, where $M$ $\in$ \{CLS, OpenGAN\},
    corresponds to a model defined on raw pixels.
  \item $M^{fea}$ corresponds to a model defined on embedding features at the penultimate layer of
    underlying semantic segmentation network (i.e., HRNet as introduced below).
  \item HRNet~\cite{WangSCJDZLMTWLX19} is a top-ranked semantic segmentation model on Cityscapes.
    It has a multiscale pyramid head that produce
    high-resolution segmentation prediction.
    We extract embedding features at its penultimate layer (720-dimensional before the 19-way classifier).
    We also tried other layers but we did not observe significant difference in their performance.
\end{itemize}

{\bf Batch Construction.}
To fully shuffle open- and closed-set training pixels,
we cache all the open-set training pixel features extracted from HRNet.
We construct a batch consisting of 10,000 pixels for training OpenGAN$^{fea}$.
To do so, we
\begin{itemize} % [noitemsep, topsep=-2pt]
  \item randomly sample a real image,
    run HRNet over it and randomly extract 5,000 closed-set training pixel features;
  \item randomly sample 2,500 open-set training features from cache;
  \item run the OpenGAN$^{fea}$ generator (being trained on-the-fly) to
    synthesize 2,500 ``fake'' open-set pixel features.
\end{itemize}
Similarly, to train OpenGAN$^{pix}$ which is fully-convolutional,
we construct a batch of 10,000 pixels as below.
\begin{itemize} % [noitemsep, topsep=-2pt]
  \item We feed a random real image to the OpenGAN$^{pix}$ discriminator, and penalize predictions on 5000 random closed pixels and 2500 random open pixels.
  \item We run the OpenGAN$^{pix}$ generator (being trained on-the-fly) to synthesize a ``fake'' image. We feed this ``fake'' image to the discriminator along with \emph{open-set} labels. We penalize 2500 random ``fake'' pixels.
\end{itemize}

\begin{figure*}[t]
\centering
\includegraphics[width=0.80\linewidth]{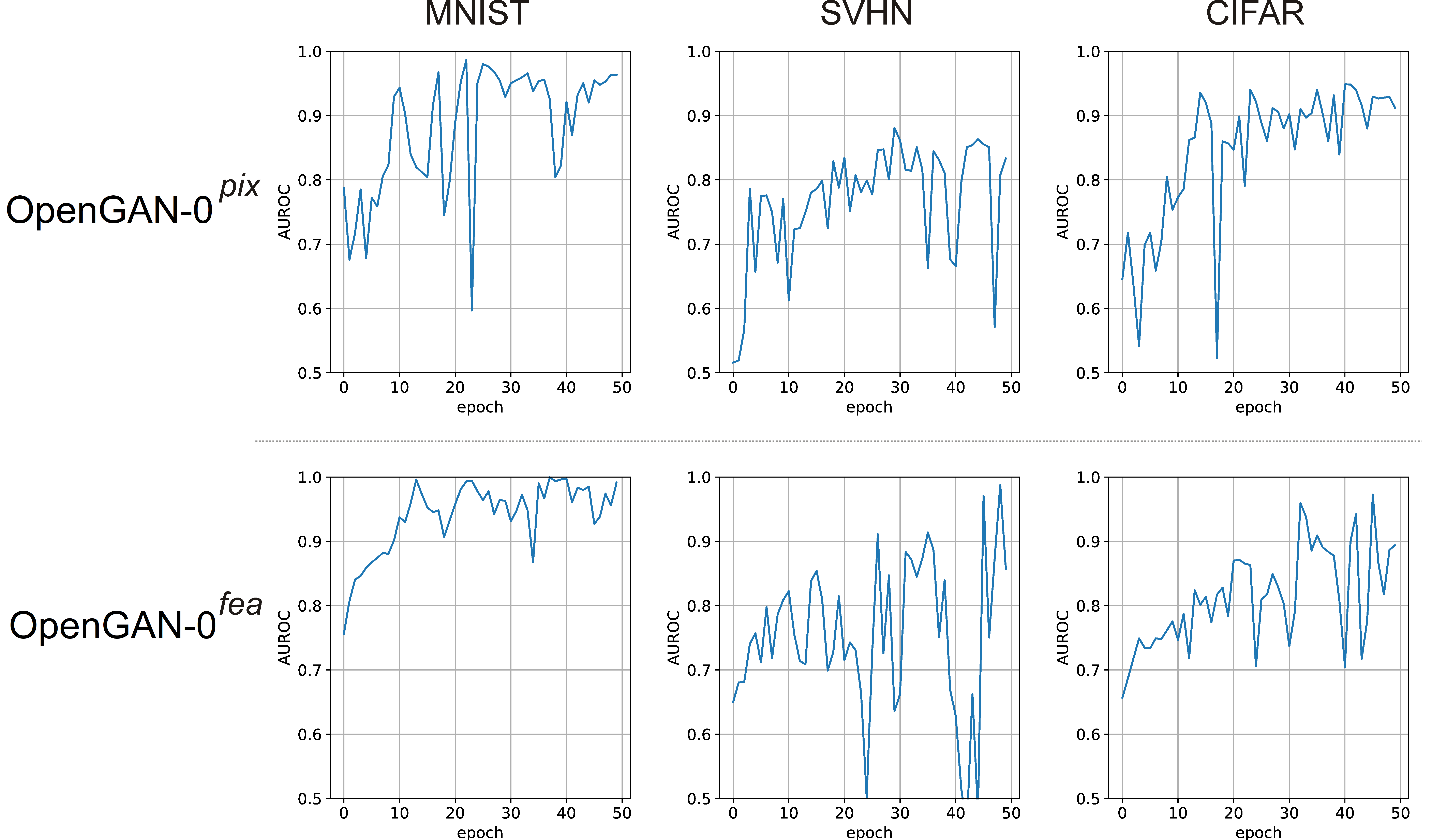}
\vspace{-3mm}
\caption{\small
{\bf Open-set image  recognition performance vs. training epochs}. We show the performance (AUROC) by OpenGAN-0$^{pix}$ and OpenGAN-0$^{fea}$ on the val-sets of the three datasets which are widely studied in the open-set recognition literature ({\em Setup-I}).
Recall that OpenGAN-0 is to train a normal GAN and use its discriminator as open-set likelihood function for open-set recognition.
We can see that best open-set discrimination performance is achieved by intermediate checkpoints of GAN discriminators, and longer training does \emph{not} necessarily improve performance. This is due to the unstable training of GANs with the min-max game.
This motivates the need for robust model selection.
}
\vspace{-3mm}
\label{fig:AUROC_vs_epoch}
\end{figure*}

\begin{figure*}[t]
\centering
   \includegraphics[width=0.8\linewidth]{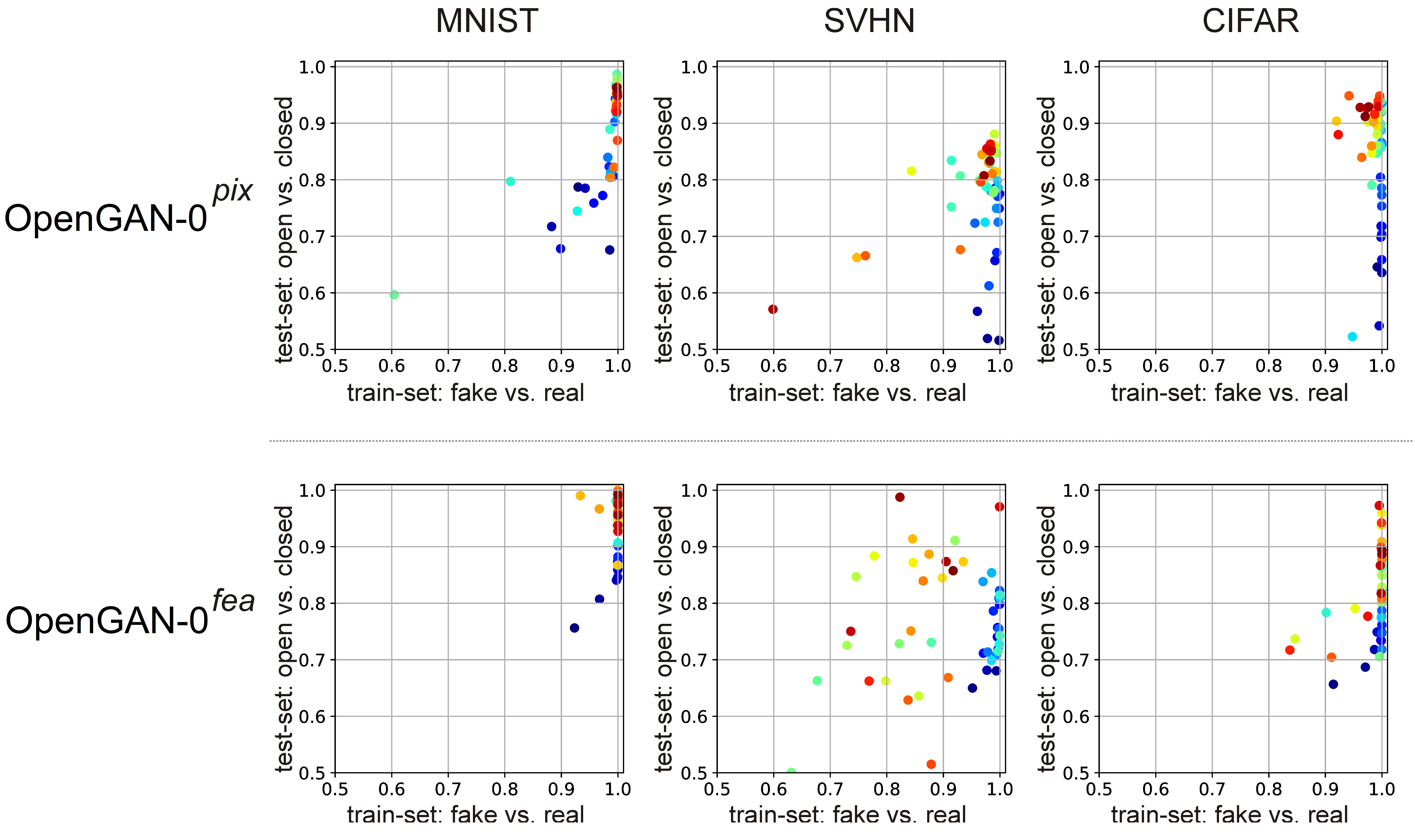}
\vspace{-4mm}
\caption{\small
{\bf Scatter plot of training performance (fake-vs-real classification) and testing performance (open-vs-closed classification).}
We color the dots from \textcolor{darkblue}{blue}$\rightarrow$ \textcolor{darkred}{red} to marks models saved at epoch-0$\rightarrow$50, respectively.
We use the three datasets (under {\em Setup-I}) following the typical setup of open-set recognition. The ideal correlation is that all the dots lie in the diagonal from bottom-left to top-right. However, there is no correlation between training (fake-vs-real) and validation (open-vs-closed) performance. Moreover, because the dots appear to be on the right part in the plots, this means that fake-vs-real classification (as denoted by the $x$-axis) is much easier than open-vs-closed classification (as denoted by the $y$-axis). These scatter plots demonstrate that (1) intermediate discriminators can perform quite well in open-set discrimination (i.e., on the validation set consisting of real open and closed-set images), and (2) synthesized data are insufficient to be used for model selection.
}
\vspace{-3mm}
\label{fig:scatter_train_vs_test}
\end{figure*}

% \newpage
\section{Model Selection}
\label{sec:model-selection}

Due to the unstable training  of GANs~\cite{arjovsky2017wasserstein}, model selection is crucial and challenging. GANs are typically used for generating realistic images, so model selection for GANs focuses on  selecting generators. To do so, one relies heavily on manual inspection of visual results over the generated images from different model epochs~\cite{goodfellow2014generative}.
In contrast, we must select the discriminator, rather than the generator, because we use the discriminator as an open-set likelihood function for open-set recognition.
It is important to note that, in theory, a perfectly trained discriminator would not be capable of recognizing fake open-set data because of the equilibrium in the discriminator/generator game~\cite{arora2017generalization}. Although such an equilibrium hardly exist in practice, we find it crucial to select GAN discriminators to be used as open-set likelihood function.
For model selection, we further find it crucial to use a validation set that consists of both real open and closed data. We present this study below.

{\bf Model selection is crucial.}
In Figure~\ref{fig:AUROC_vs_epoch}, we plot the open-set classification performance as a function of training epochs. We study both OpenGAN-0$^{fea}$ and OpenGAN-0$^{pix}$ on the three datasets as typically used in open-set recognition (under {\em Setup-I}).
Recall that OpenGAN-0 is to train a normal GAN and use its discriminator as open-set likelihood function for open-set recognition.
Clearly, we can see that long training time does \emph{not} necessarily improve open-set classification performance. We posit that this is due to the unstable training of GANs. This motivates robust model selection using a validation set.

{\bf Synthesized data are not sufficient for model selection.}
To study how each checkpoint models perform in training (fake-vs-real classification) and testing (open-vs-closed classification), we scatter-plot   Figure~\ref{fig:scatter_train_vs_test}, where we render the dots with colors to indicate the model epoch (\textcolor{darkblue}{blue}$\rightarrow$ \textcolor{darkred}{red} dots represent model epoch-0$\rightarrow$50, respectively).
For the scatter plot, the ideal case is that the  train-time and test-time performance is linearly correlated, i.e., all dots appear in the diagonal line (from origin to top right). But their performances on the two sets are not correlated, suggesting that using the synthesized data for model selection is not sufficient. Instead, we find it crucial to use a validation set of real open examples to select the  open-set discriminator. Our observation is consistent to what reported in~\cite{hendrycks2018deep}.
It is worth noting that the models selected on the validation set \emph{do} generalize to test sets. This has been demonstrated in Table 3 and 4 in the main paper.

\section{Hyper-Parameter Tuning}
\label{sec:param-tuning}

{
\setlength\tabcolsep{7.5pt}
\begin{table*}
\centering
\captionof{table}{\small
{\bf Hyper-parameter tuning for open-set semantic segmentation  on Cityscapes.} Given a fixed number of open training images, we vary the hyper-parameter $\lambda_G$ to train OpenGAN models. Recall that $\lambda_G$ controls the contribution of synthesized data in the loss function. We conduct model selection on the val-set (10 images), and report here the performance (AUROC$\uparrow$) on the test set (500 images). We also mark the $\lambda_G$ for each of the selected models. It seems to be preferable to set a lower weight $\lambda_G$ (for the term exploiting synthesized data examples) when we have more real open-set data, but we do not see a tight correlation between $\lambda_G$ and test-time performance. We believe this is because of the (random) initialization of model weights that has a non-trivial impact on training GANs and their final performance.}
\begin{tabular}{l| c c c c c c c c c}
\hline
$\#$images$_{train}^{open}$
& {\tt 10} & {\tt 20}  & {\tt 50} & {\tt 100}  & {\tt 200} & {\tt 500} & {\tt 1000} & {\tt 2000} & {\tt 2900} \\
\hline
OpenGAN$^{fea}$  & .761  & .821 & .849 & .866 & .891 &.890 & .873 & .891 & .885 \\
$\lambda_G$  & 0.20  & 0.20 & 0.05 & 0.10 & 0.05 & 0.05 & 0.20 & 0.05 & 0.05 \\
\hline\hline
OpenGAN$^{pix}$  & .607  & .632 & .643 & .661 & .672 &.705 & .711 & .748 & .746 \\
$\lambda_G$  & 0.60  & 0.40 & 0.80 & 0.90 & 0.70 & 0.60 & 0.60 & 0.60 & 0.70 \\
\hline
\end{tabular}
\label{tab:param_tuning}
\end{table*}
}

\begin{figure}[t]
\centering
   \includegraphics[width=1\linewidth]{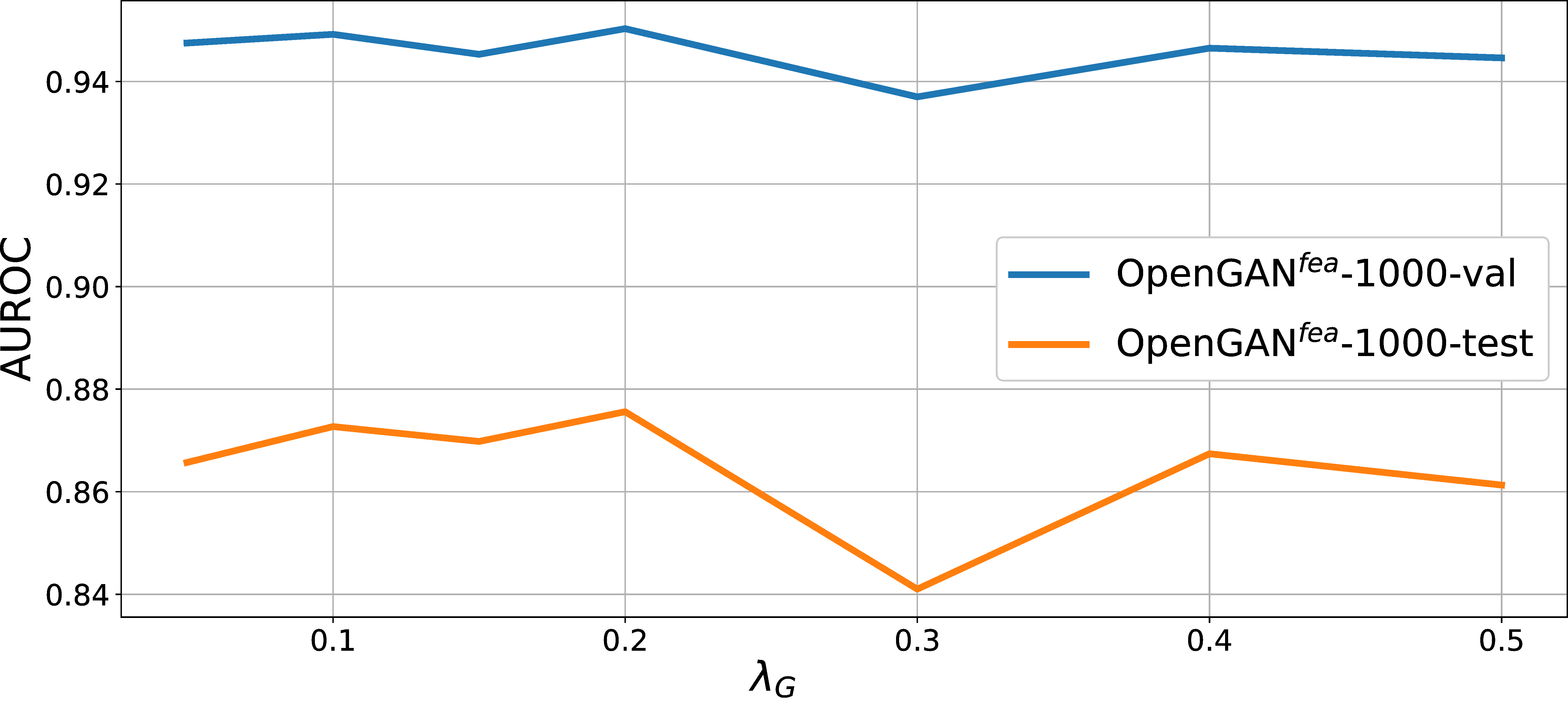}
\vspace{-3mm}
\caption{\small
{\bf Tuning hyper-parameter $\lambda_G$.} We plot the open-set discrimination performance (AUROC) as a function of $\lambda_G$, which controls the contribution of generated data examples in the loss function. The model we report here is OpenGAN$^{fea}$-1000 that is trained with 1000 open-set training images. The validation set and test set contain 10 and 500 images. Although the validation set has much fewer images than the test set, the open-set classification performances align well on the two sets.
}
\vspace{-2mm}
\label{fig:roc_vs_lambda}
\end{figure}

\begin{figure}[t]
\centering
\includegraphics[width=1\linewidth]{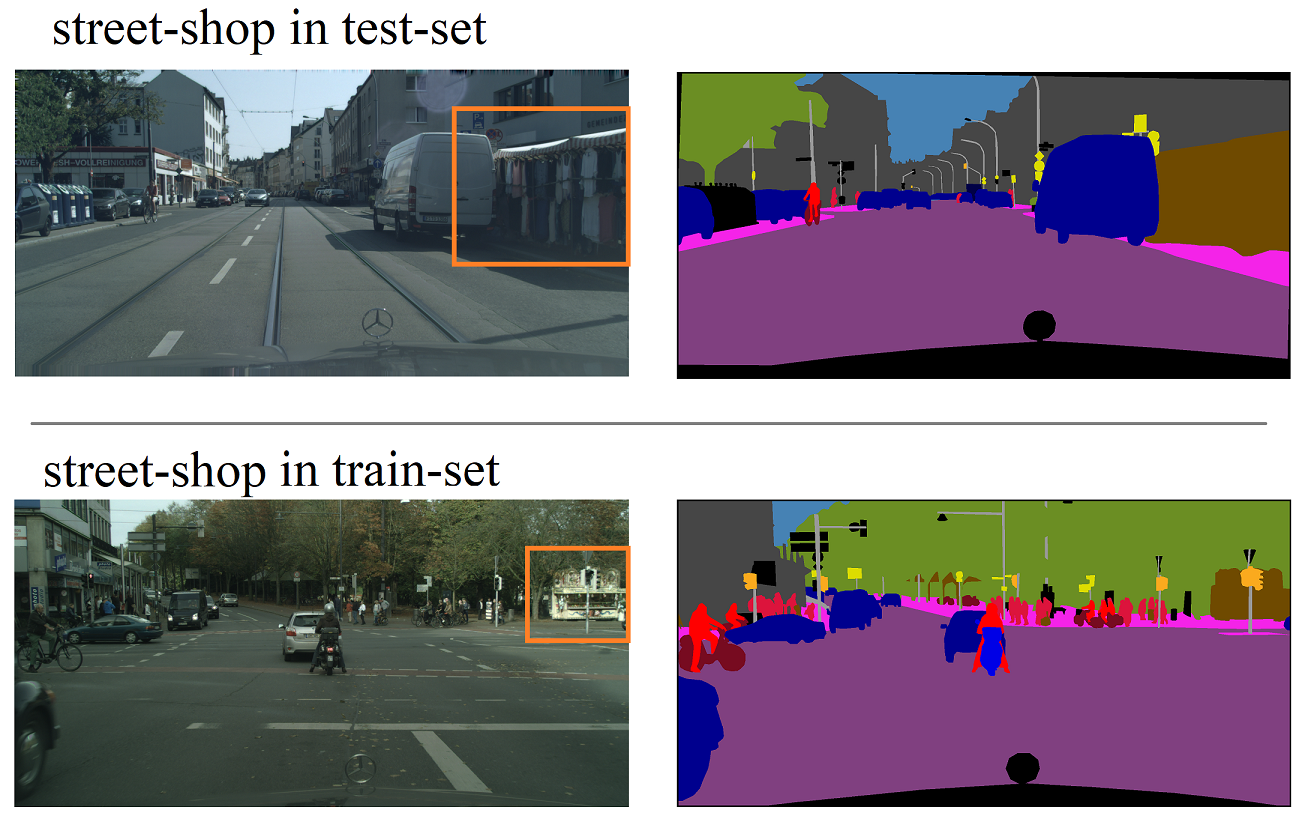}
\vspace{-3mm}
\caption{\small
{\bf street-shop as open-set.}
Figure 4 in the main paper shows open-set pixel recognition results on a street-shop on a testing image (top-row). We verify if such a street-shop appears in the training set.
We manually search for a similar street-shop in the training set, and find the one (bottom-row) most similar to the testing example in terms of size. Importantly, we did not find any other street-shops in the training set that sell clothes like the  testing example shown in the top row. In this sense, the testing image in the top row does contain a real open-set example (i.e., the street-shop) in terms of not only size, but also novel content.
}
\vspace{-2mm}
\label{fig:street-shop}
\end{figure}

Strictly following the practice of machine learning, we tune hyper-parameters on the same validation set.
We now study parameter tuning through open-set semantic segmentation ({\em Setup-III}). We select the best OpenGAN model according to the performance on the validation set (10 images).

In training OpenGAN, a training batch contains both real closed- and open-set pixels, and synthesized \emph{fake open} pixels. Correspondingly, our loss function has three terms (refer to Eq. 2 in the main paper). Therefore, we tune the hyper-parameter $\lambda_o$ and $\lambda_G$ as below to balance the terms in the loss function that exploits real open data and generated data:
\begin{itemize}
  \item The term exploiting real open data has a weight $\lambda_o=1$. We do not tune this as we presume the sparsely sampled open-set examples are equally important as the real closed-set examples.
  \item The term using the generated ``fake'' data has varied parameter $\lambda_G \in$ [0.05, 0.10, 0.15, 0.20, $\dots$, 0.80, 0.85, 0.90]. We mainly focus on tuning $\lambda_G$ to study how the synthesized data help training.
\end{itemize}

In Table~\ref{tab:param_tuning},  we show the performance on the test set of OpenGAN$^{pix}$ and OpenGAN$^{fea}$ with varied open training images. For each selected model, we mark the corresponding $\lambda_G$ that yields the best performance (on validation set).
Roughly speaking,
it is preferable to set a lower weight $\lambda_G$  when we have more real open-set training data.
However, we do not see a clear correlation between the weight $\lambda_G$ and test-time performance. We believe this is due to the random initialization which affects adversarial learning.

We also study how models trained with different $\lambda_G$ perform on validation set and test set, and if the model selected on the validation set can reliably perform well on the testing set.
Figure~\ref{fig:roc_vs_lambda} plots the performance as a function of $\lambda_G$ on validation set and test set.
Hereby we choose the OpenGAN$^{fea}$ model trained with 1000 open training images.
We can see the performance on the validation set reliably reflects the performance on the test set. This confirms that model selection on the validation set is reliable.

\section{Statistical Models for Open-Set}
\label{sec:GMM}

\begin{figure*}[t]
\centering
\centering
\includegraphics[width=1\linewidth]{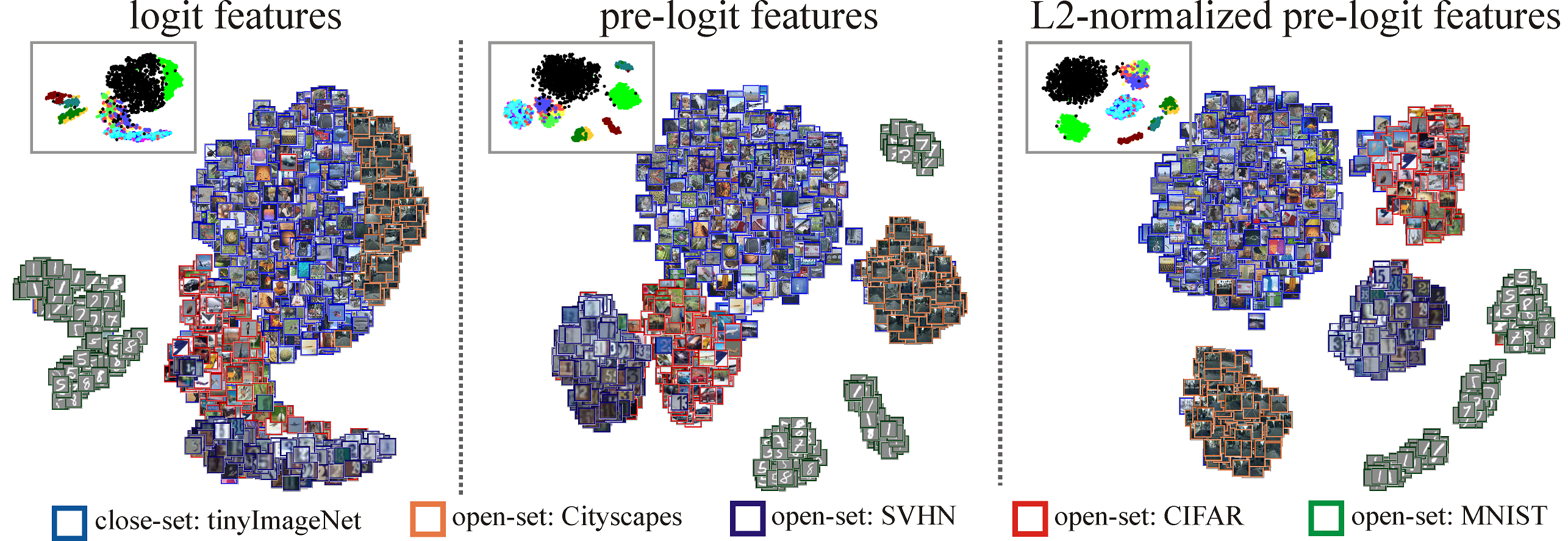}
\vspace{-2.5mm}
\caption{\small
  t-SNE plots~\cite{maaten2008visualizing} of open vs closed-set testing data, as encoded by different features from a ResNet18 network trained from scratch on the TinyImageNet dataset for 200-way classification. To better view the clustering results, we show  zoom-out scatter plots in which we show closed-set data in black, and color open-set examples using their class labels provided by the respective datasets.
  {\bf Left}: Logit features mix open and closed data, suggesting that methods based on them (Entropy, SoftMax and OpenMax) may struggle in open-set discrimination.
  {\bf Mid}: Pre-logit features at the penultimate layer show better separation between closed- and open-set data.
  {\bf Right}: Normalizing the pre-logits features separates them even better. These plots intuitively demonstrate the benefit of L2-normalization and using OTS features rather than the highly-invariant logits.}
\vspace{-1mm}
\label{fig:why_ft}
\end{figure*}

Our previous work introduces a lightweight statistical pipeline that repurposes off-the-shef (OTS) deep features for open-set recognition~\cite{kong2021an}. For the completeness of this paper, we briefly introduce this pipeline: (1) extracting OTS features (with appropriate processing detailed below) of closed-set training examples using the underlying $K$-way classification model, (2) learning statistical models over the OTS features. There are many statistical methods one can choose, e.g., nearest class centroids, Nearest Neighbors, and (class-conditional) Gaussian Mixture Models (GMMs).
During testing, we extract the OTS features of the given example and resort to the learned statistical models to compute an open-set likelihood, e.g., based on the (inverse) closed-set probability from GMM. By thresholding the open-set likelihood, we decide whether it is an open-set example or one of the $K$ closed-set classes, with the latter we report the predicted class label.

{\bf Feature extraction}.
OTS features generated at different layers of the trained $K$-way classification network can be repurposed for open-set recognition. Most methods leverage softmax~\cite{hendrycks2016baseline} and logits~\cite{bendale2016towards, grathwohl2019your, OzaP19} which can be thought of as features extracted at top layers.
Similar to \cite{lee2018simple}, we find it crucial to analyze features from intermediate layers for open-set recognition, because logits and softmax may be too invariant to be effective for open-set recognition (Fig.~\ref{fig:why_ft}). One immediate challenge to extract features from an intermediate layer is their high dimensionality, e.g. of size 512x7x7 from ResNet18~\cite{he2016deep}. To reduce feature dimension, we simply (max or average) pool the feature activations spatially into a 512-dim feature vectors~\cite{yang2015multi}.
We can further reduce dimension by apploying PCA, which can reduce dimensionality by 10$\times$ (from 512-dimensional to 50 dimensional) without sacrificing performance. We find this dimensionality particularly important for learning second-order covariance statistics as in GMM, described below.
Finally, following~\cite{gong2014multi, gordo2017end},
we find it crucial to L2-normalize extracted features (Fig.~\ref{fig:why_ft}). We refer the reader to \cite{kong2021an} for quantitative results.
Note in this paper, we do not L2-normalize features for training OpenGANs.

{\bf Statistical models.}
Given the above extracted features, we use various generative statistical methods to learn the confidence/probability that a test example belongs to the  closed-set classes. Such statistical methods include  simple parametric models such as class centroids~\cite{mensink2012metric} and class-conditional Gaussian models~\cite{lee2018simple, grathwohl2019your},  non-parametric models such as NN~\cite{boiman2008defense, junior2017nearest}, and mixture models such as  (class-conditional) GMMs and k-means~\cite{cao2016deep}.
A statistical model labels a test example as open-set when the inverse probability (e.g., of the most-likely class-conditional GMM) or distance (e.g., to the closest class centroid) is above a threshold.
One benefit of such simple statistical models is that they are interpretable and relatively easier to diagnose failures. For example, one failure mode is an open-set sample being misclassified as a closed-set class. This happens when open-set data lie close to a class-centroid or Gaussian component mean (see Fig.~\ref{fig:why_ft}).
Note that a single statistical model may have several hyperparameters --  GMM can have multiple Gaussian components and different structures of second-order covariance, e.g., either a single scalar,  a diagonal matrix or a general covariance per component.
We make use of validation set to determine these hyperparameters,
as opposed to prior works that conduct model selection either unrealistically on the test-set~\cite{OzaP19} or on large-scale val-set which could be arguably used for training~\cite{lee2018simple}.
We refer the reader to \cite{kong2021an} for detailed analysis.

{\bf Lightweight Pipeline}.
We re-iterate that the above feature extraction and statistical models result in a lightweight pipeline for open-set recognition.
To understand this, we analyze the number of parameters involved in the pipeline. Assume we learn a GMM over 512x7x7 feature activations,
and specify a general covariance and five Gaussian components.
If we learn the GMM directly on the feature activations,
the number of parameters from the second-order covariance alone is
at the scale of $(512*7*7)^2$.
With the help of our feature extraction (including spatial pooling and PCA),
we have 50-dim feature vectors,
and the number of parameters in the covariance matrices is now at the scale of $50^2$.
This means a huge reduction ($10^5$ $\times$) in space usage!
We count the total number of parameters in this GMM:
3.3$\times$ $10^{4}$ 32-bit float parameters including PCA and GMM's five components,
amounting to 128KB storage space.
Moreover,
given that PCA just runs once for all classes,
even when we learn such GMMs for each of 19 classes (such as defined in Cityscapes), it merely requires 594KB storage space!
Compared to the modern networks such as HRNet ($>$250MB),
our statistical pipeline for open-set recognition adds a negligible (0.2\%) amount of compute, making it quite practical for implementation on autonomy stacks.

\section{Further Quantitative Results}
\label{sec:sota-comparison}
While in the main paper we compare OpenGAN to many methods and cannot include more due to space issues,
we list a few more in this appendix including  Entropy~\cite{steinhardt2016unsupervised}, GMM~\cite{kong2021an}, CGDL~\cite{sun2020conditional}, OpenHybrid~\cite{zhang2020hybrid},  and RPL++~\cite{chen2020learning}. Except for Entropy which is a classic method, the rest were published recently.
Table~\ref{tab:toy_openset} lists the comparisons under {\em Setup-I}. Please refer to Section 4.2 of the main paper for the detailed setup. Numbers are comparable to Table 1 in the main paper.
In summary, our OpenGAN outperforms all these prior methods under this setup, achieving the state-of-the-art.

{
\setlength{\tabcolsep}{0.1em} % for the horizontal padding
\begin{table*}[t]
% \scriptsize
%\footnotesize
\small
\centering
\caption[Caption for LOF]{\small
{\bf Open-set discrimination (Setup-I)} measured by  area under ROC curve (AUROC)$\uparrow$. Numbers are comparable to Table 1 in the main paper.
Recall that OpenGAN-0 does not train on outlier data (i.e., $\lambda_0$=0 in Eq. 2) and only selects discriminator checkpoints on the validation set. OpenGAN-0$^{fea}$ clearly performs the best, achieving the state-of-the-art.
}
\vspace{-3mm}
\begin{tabular}{l | c  c  c  c  c  c  c  c  c  c  c  c  c | c }
%\toprule
\hline
 & MSP & Entropy & OpenMax & MSP$_{c}$ & GOpenMax & OSRCI & MCdrop & GDM & GMM & C2AE & CGDL & RPL-WRN & OpenHybrid & OpenGAN-0$^{fea}$  \\
Dataset &  \cite{hendrycks2016baseline}  & \cite{steinhardt2016unsupervised}  & \cite{bendale2016towards}
& \cite{liang2017enhancing}
&  \cite{ge2017generative}  &  \cite{neal2018open}
& \cite{gal2016dropout}
& \cite{lee2018simple}
& \cite{kong2021an}  & \cite{OzaP19}
& \cite{sun2020conditional}
&  \cite{chen2020learning}
& \cite{zhang2020hybrid}
& ({\em ours}) \\
\hline
\emph{MNIST}  & .977  & .988  & .981
& .985
& \textcolor{black}{.984}  & \textcolor{black}{.988}
& .984
& .989
& .993 & .989
& .994
& .996
& .995
& \textbf{.999} \\
\hline
\emph{SVHN}   &  .886 & .895  & .894
& .891
& \textcolor{black}{.896}  & \textcolor{black}{.910}
& .884
& .866
& .914 & .922
& .935
& .968
& .947
& {\bf .988} \\
\hline
\emph{CIFAR}& .757  & .788  & .811
& .808
& \textcolor{black}{.675}  & \textcolor{black}{.699}
& .732
& .752
& .817 & .895  % .895 is what reported in their paper, .801 is from our re-implementation
& .903
& .901
& .950
& {\bf .973} \\
%\bottomrule
\hline
\end{tabular}
\label{tab:toy_openset}
\vspace{-3mm}
\end{table*}
}

\section{Visualization of Generated Images}
\label{sec:visualization}

\begin{figure*}[t]
\centering
\includegraphics[width=0.90\linewidth]{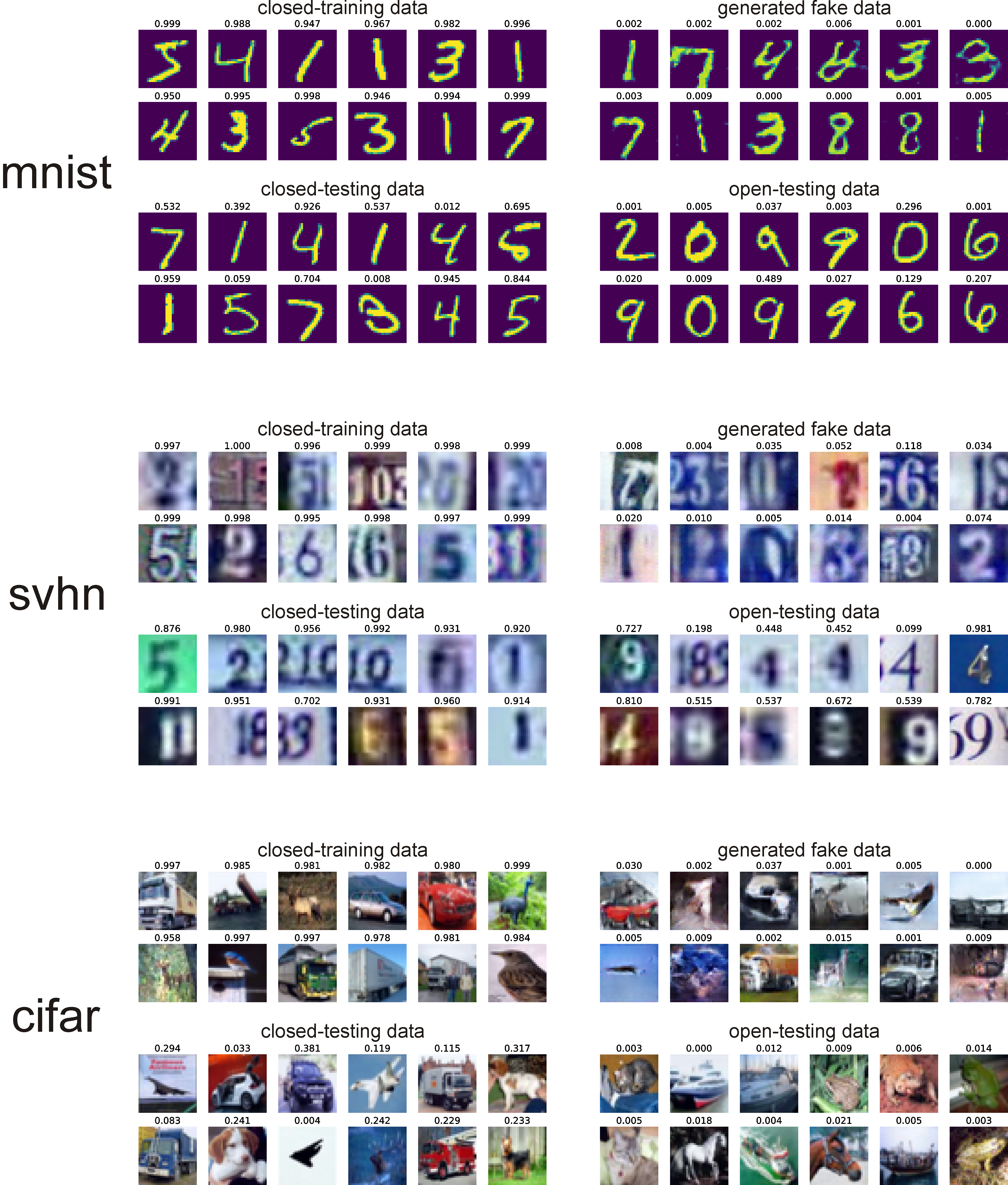}
%\vspace*{-3mm}
\caption{\small
{\bf Demonstration of visuals along with the classification confidence scores as probabilities of being recognized as closed-set data.}
On each of the three datasets,
we show some random images that are closed-training image (for training GANs), synthesized ``fake'' images, closed-testing images (from known classes) and open-testing images (from unknown classes). We also mark the probability for each image of being classified as closed-set  by the  discriminator. We can see the synthesized images look realistic in terms of color, tone and shape. But the discriminator can easily recognize these fake images (as indicated by the low probability). Moreover, although the discriminator achieves good open-vs-closed classification performance measured by AUROC (which is calibration-free), the confidence scores (probability) are not calibrated well. This implies that the discriminator may need to be calibrated for real-world application.
}
%\vspace{-5mm}
\label{fig:viz_toy_example}
\end{figure*}

\begin{figure*}[t]
\centering
\includegraphics[width=0.99\linewidth]{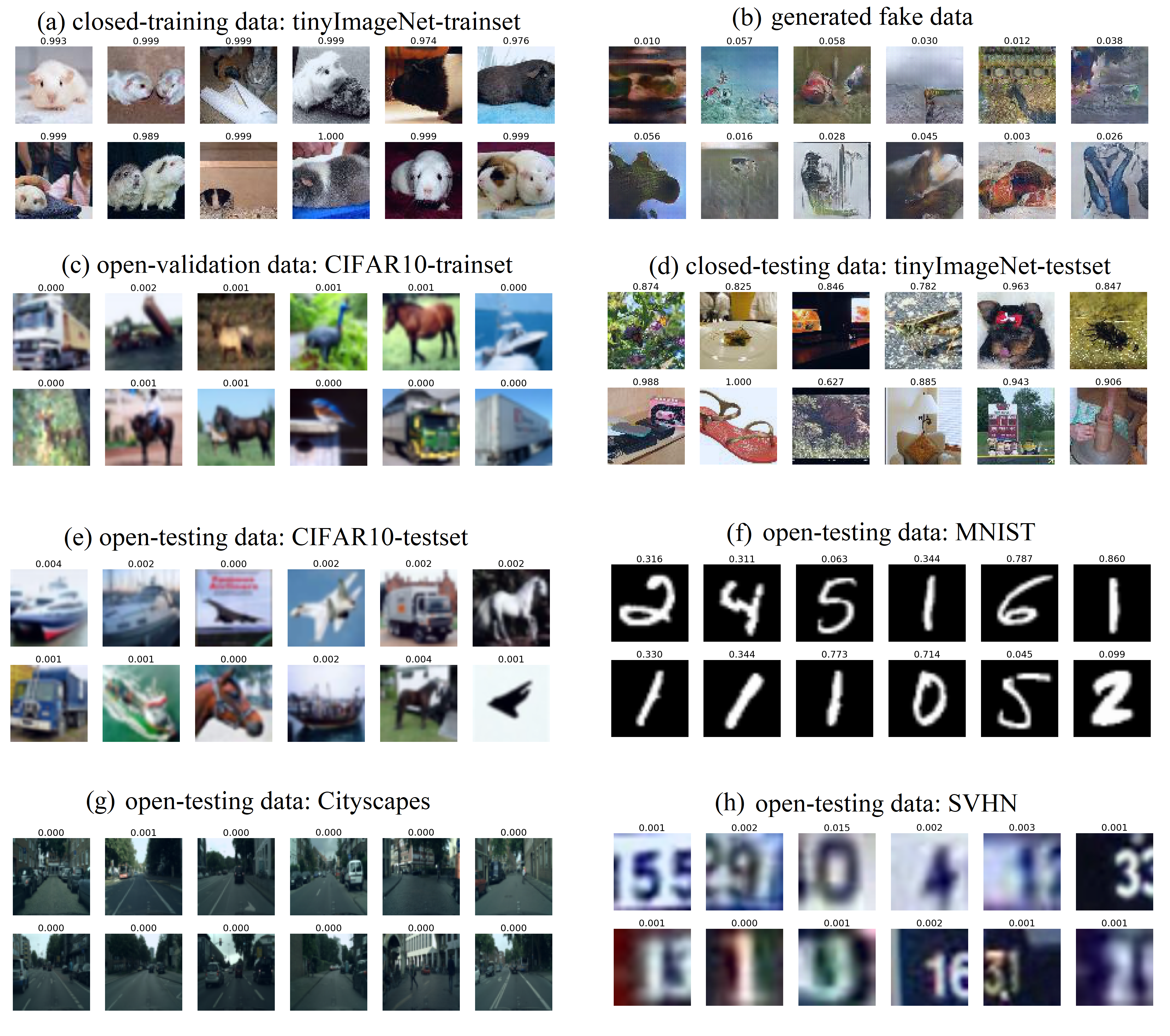}
%\vspace*{-3mm}
\caption{\small
{\bf Demonstration of visuals along with the OpenGAN-0$^{pix}$ classification confidence scores as probabilities of being recognized as closed-set data.}
These visual results are generated under \emph{Setup-II}, where the TinyImageNet is the closed-set for 200-way classification, and other datasets are treated as the open-set.
The discriminator of the OpenGAN-0$^{pix}$ is seleced over the CIFAR train-set.
{\bf (a)} The discriminator recognizes the closed-set training examples with a high confidence score.
{\bf (b)} OpenGAN-0$^{pix}$ synthesizes fake images that look realistic in terms of color, tone and shape, but not content.
The discriminator can easily recognize these fake images (as indicated by the low probability).
The discriminator generalizes well in terms of recognizing closed-set examples from the validation and test sets as shown in {\bf (c)} and {\bf (d)}, and open-set examples from other datasets as shown in {\bf (e)}, {\bf (f)}, and {\bf (h)}.
}
%\vspace{-5mm}
\label{fig:viz_tinyImageNet}
\end{figure*}

\begin{figure*}[t]
\centering  %\fbox{\rule{0pt}{2in} \rule{0.9\linewidth}{0pt}}
\hspace{6mm} Real \hspace{34mm}  OpenGAN-0$^{pix}$ \hspace{28mm} OpenGAN-0$^{fea}$\\
\includegraphics[width=0.85\linewidth, clip = true, trim = 0mm 0mm 0mm 34mm]{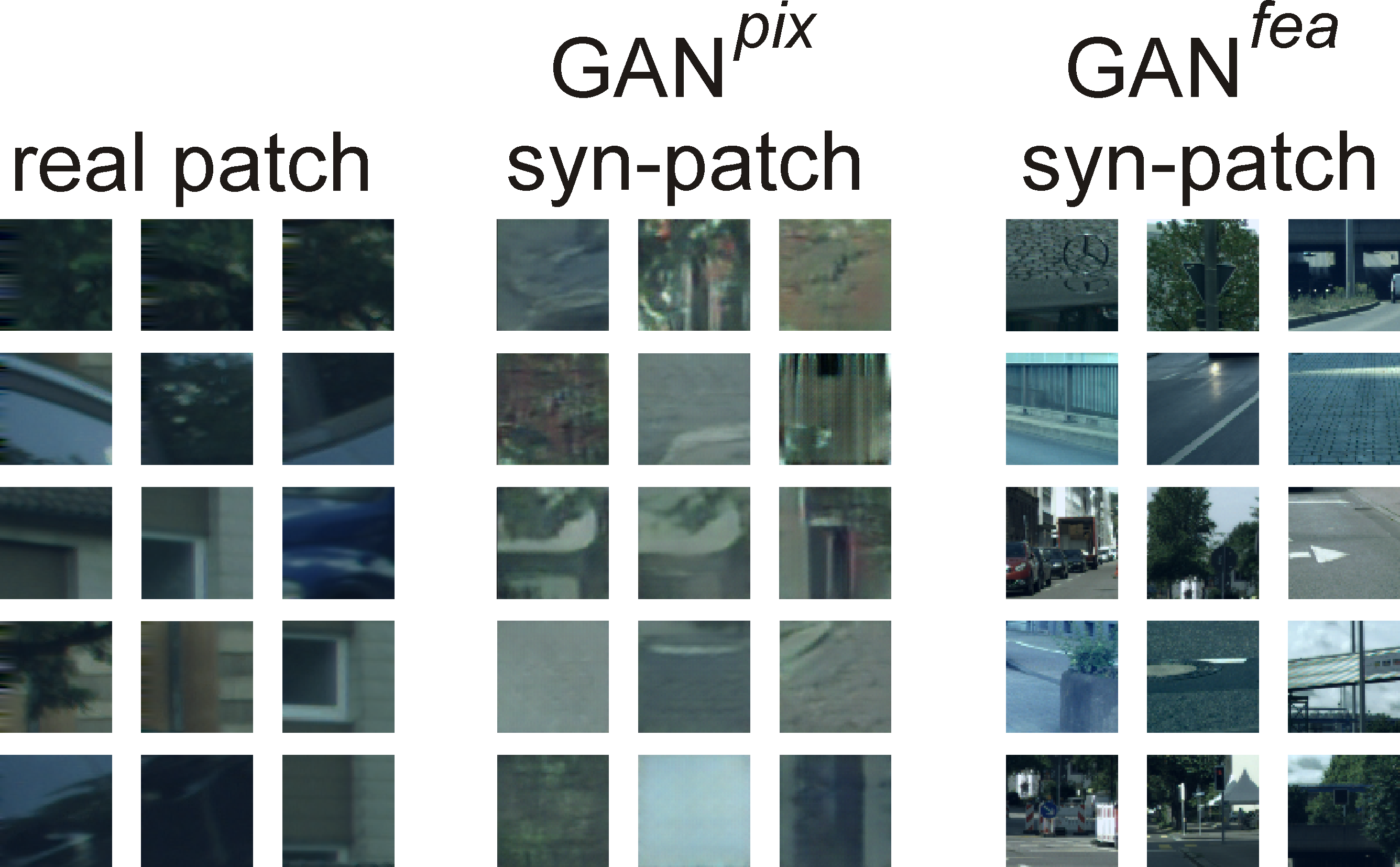}
%\vspace*{-3mm}
\caption{\small
{\bf Visuals of real Cityscapes image patches (left), synthesized patches by OpenGAN-0$^{pix}$ (mid) and GOpenGAN-0$^{fea}$ (right).}
As OpenGAN-0$^{fea}$ generates feature vectors instead of RGB patches, we ``synthesize'' the patches in an analytical way -- for a generated feature, we find the nearest-neighbor per-pixel feature (w.r.t L1 distance) from the training images, and then find the RGB patch centered at the associated  pixel with the per-pixel feature.  The real patch is our ``synthesized'' patch for that generated feature.
The synthesized patches by OpenGAN-0$^{pix}$ do look realistic in terms of color and tone, but OpenGAN-0$^{pix}$ (0.549 AUROC) does not work as well as OpenGAN-0$^{fea}$ (0.709 AUROC).
The ``synthesized'' patches by OpenGAN-0$^{fea}$ do capture some \emph{unknown open-set} objects, such as {\tt bridge}, {\tt back of traffic sign} and {\tt unknown static objects}, none of which belong to any of the 19 classes defined in Cityscapes for semantic segmentation (cf. Figure~\ref{fig:histogram_in_cityscapes}).
}
%\vspace{-5mm}
\label{fig:genImages_cityscapes}
\end{figure*}

In this section, we visualize some synthesized examples for intuitive demonstration.

{\bf Generating Small Images.}
Recall that OpenGAN-0$^{pix}$ trains a normal GAN and uses its discriminator as the open-set likelihood function.
As demonstrated in the main paper, OpenGAN-0$^{pix}$  performs surprisingly well under {\em Setup-I} (i.e., using CIFAR10, MNIST and SVHN datasets) and {\em Setup-II} (using TinyImageNet as the closed-set and other datasets as the open-set).
%(on-par or better than prior work).
OpenGAN-0$^{pix}$ also enables us to generate visual results for intuitive inspection.
In Figure~\ref{fig:viz_toy_example}, we display real and  synthesized ``fake'' images under \emph{Setup-I} on each of the three datasets.
In Figure~\ref{fig:viz_tinyImageNet}, we display real and fake images under \emph{Setup-II} by using tinyImageNet as the closed-set and other datasets as the open-set.
We can see the generated images look realistic in terms of color and tone. But they are not strictly open-set images as they contain synthesized \emph{known} contains (e.g., the digits in the synthesized images are closed-set digits). This intuitively demonstrates that a perfectly trained discriminator will not be capable of discriminating open and closed-sets due to the nature of the min-max game in training GANs.
However, from the low confidence scores of classifying the generated fake data as closed-set shown in Figure~\ref{fig:viz_toy_example},
we can see the discriminator almost naively recognizes these synthesized examples as ``fake'' data. This shows the synthesized data are insufficient to be used for model selection.
Moreover, from the classification confidence scores on the closed-testing and open-testing images in each datasets, we can see the discriminator is not calibrated. In other words, we cannot naively set threshold as 0.5 for open-vs-closed classification. This is largely hidden by AUROC metric which is calibration-free. This implies a potential limitation and suggests future work to calibrate the open-set discriminators.

{\bf Generating Cityscapes Patches.}
In the main paper (Fig. 6), we have shown some generated patches. In the appendix, we provide more in Figure~\ref{fig:genImages_cityscapes}.
As OpenGAN-0$^{fea}$ generates features instead of pixel patches, we ``synthesize'' the patches  analytically -- for a generated feature, from training pixels represented as OTS features, we find the nearest-neighbor pixel feature (w.r.t L1 distance), and use the RGB patch centered at that pixel as the ``synthesized'' patch. We can see OpenGAN-0$^{pix}$ synthesizes realistic patches w.r.t color and tone, but it (0.549 AUROC) significantly unperforms OpenGAN-0$^{fea}$ (0.709 AUROC) for open-set segmentation.
The ``synthesized'' patches by OpenGAN-0$^{fea}$ capture many \emph{open-set} objects, such as {\tt bridges}, {\tt vehicle logo} and {\tt back of traffic sign}, all of which are outside the 19 classes defined in Cityscapes. This intuitively explains why OpenGAN-0$^{fea}$ (0.709 AUROC) works much better than OpenGAN-0$^{pix}$ (0.549 AUROC).

% \newpage
\section{Visual Results of Open-Set Segmentation}
\label{sec:open-pixel-vis}

\begin{figure*}[t]
\begin{center}
   \includegraphics[width=0.75\linewidth]{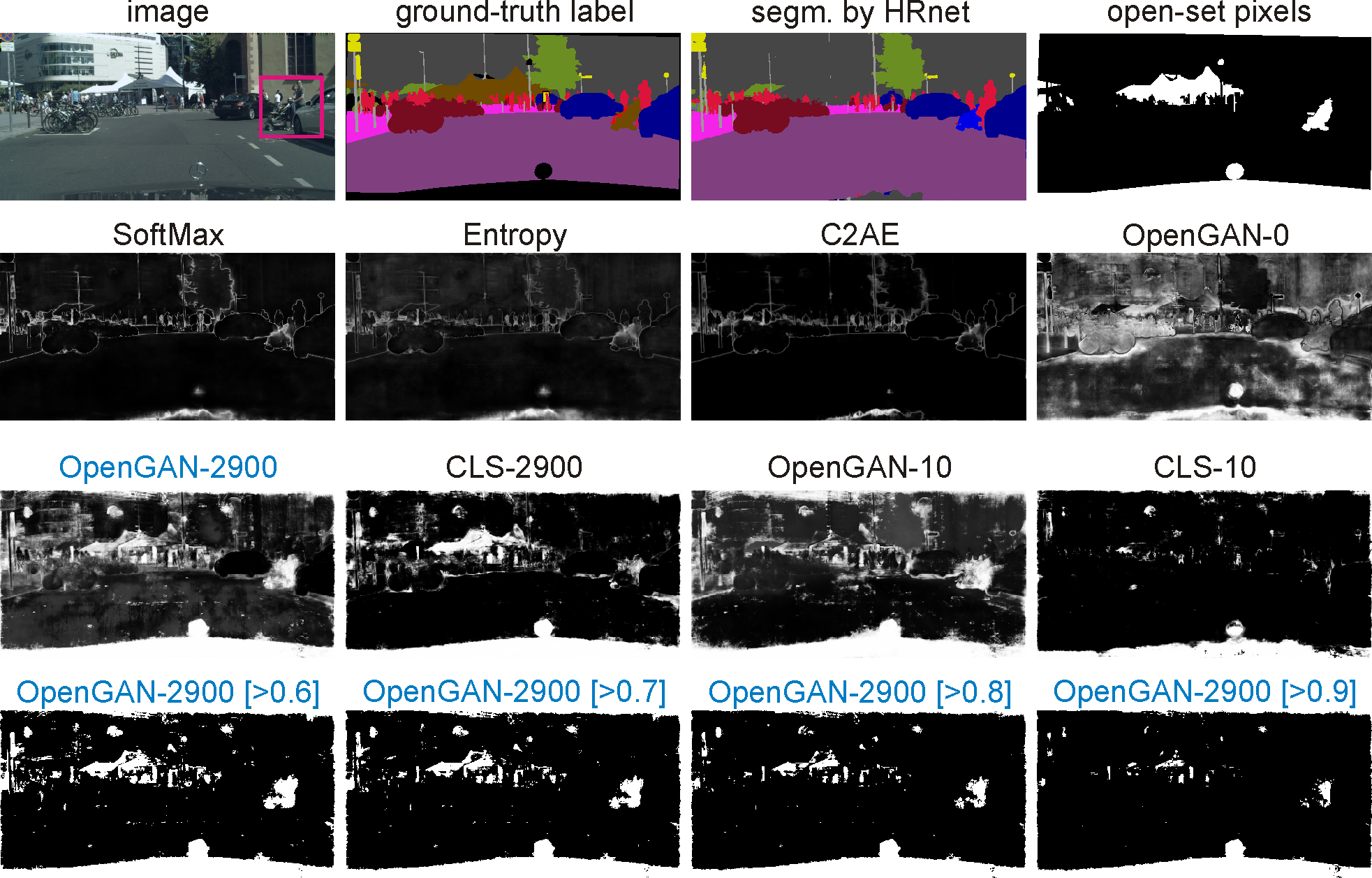}
\end{center}
\vspace*{-3mm}
\caption{\small
{\bf Qualitative results of a testing image from Cityscapes}.
[1$^{st}$ row] the input image, its per-pixel semantic labels,
the semantic segmentation result by HRnet and open-set pixels colored by white.
[2$^{nd}$ row] visual results as per-pixel scores of being classified as open-set pixel by SoftMax, Entropy, C2AE
and our OpenGAN-0$^{fea}$.
[3$^{rd}$ row] visual results by our OpenGAN$^{fea}$ and CLS,
trained with 2900 or 10 open training images, respectively.
[4$^{th}$ row] visual results by thresholding OpenGAN-2900 with 0.6, 0.7. 0.8 and 0.9 respectively.
OpenGAN clearly captures most open-set pixels (cf. the white pixels in top-right open-set map).
}
%\vspace{-5mm}
\label{fig:qualitativeDemo_appendix_136}
\end{figure*}

\begin{figure*}[t]
\begin{center}
   \includegraphics[width=0.75\linewidth]{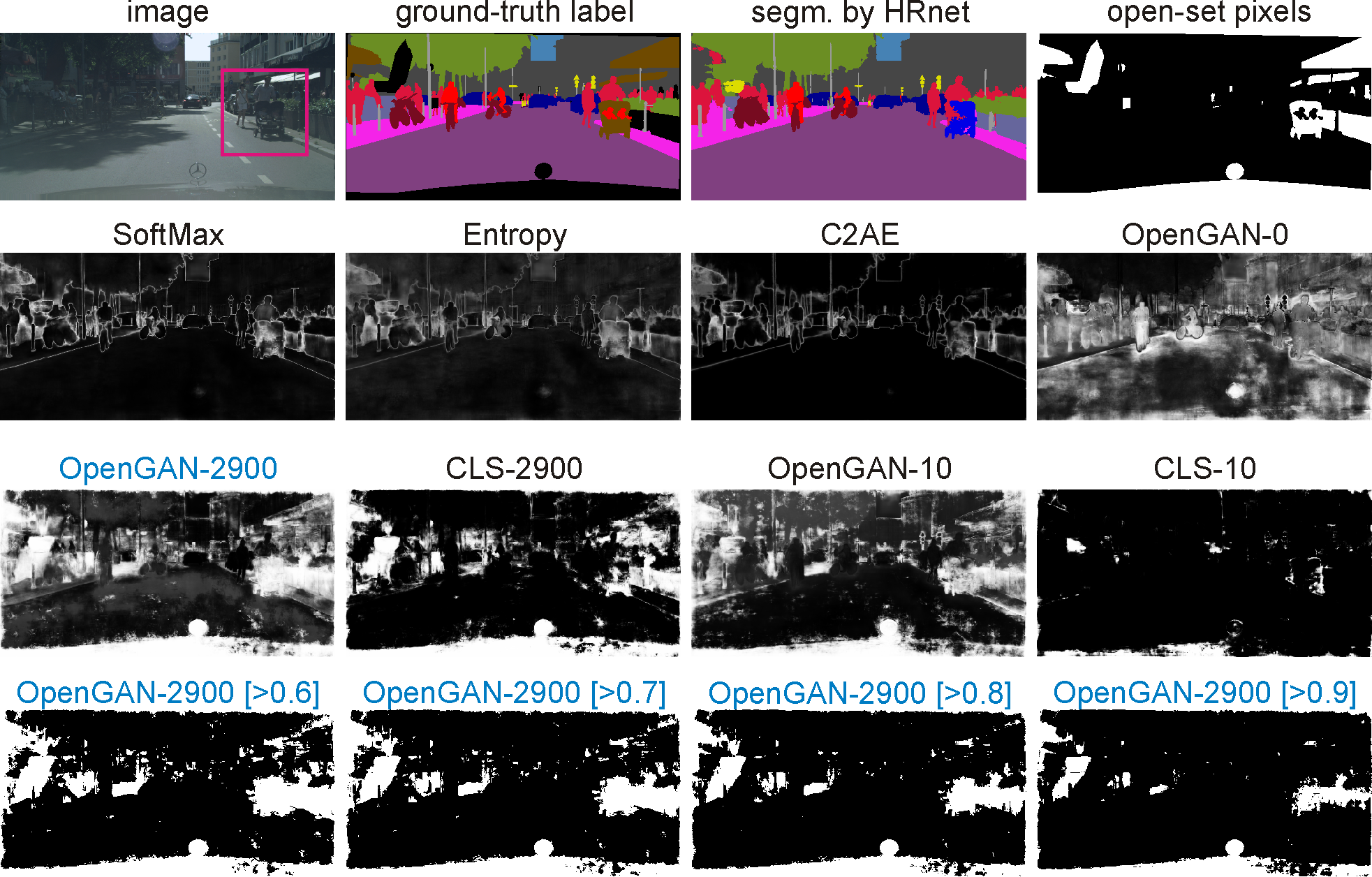}
\end{center}\vspace*{-3mm}
\caption{\small
{\bf Qualitative results of a testing image from Cityscapes}.
[1$^{st}$ row] the input image, its per-pixel semantic labels,
the semantic segmentation result by HRnet and open-set pixels colored by white.
[2$^{nd}$ row] visual results as per-pixel scores of being classified as open-set pixel by SoftMax, Entropy, C2AE
and our OpenGAN-0$^{fea}$.
[3$^{rd}$ row] visual results by our OpenGAN$^{fea}$ and CLS,
trained with 2900 or 10 open training images, respectively.
[4$^{th}$ row] visual results by thresholding OpenGAN-2900 with 0.6, 0.7. 0.8 and 0.9 respectively.
OpenGAN clearly captures most open-set pixels (cf. the white pixels in top-right open-set map).
}
%\vspace{-5mm}
\label{fig:qualitativeDemo_appendix_130}
\end{figure*}

\begin{figure*}[t]
\begin{center}
   \includegraphics[width=0.75\linewidth]{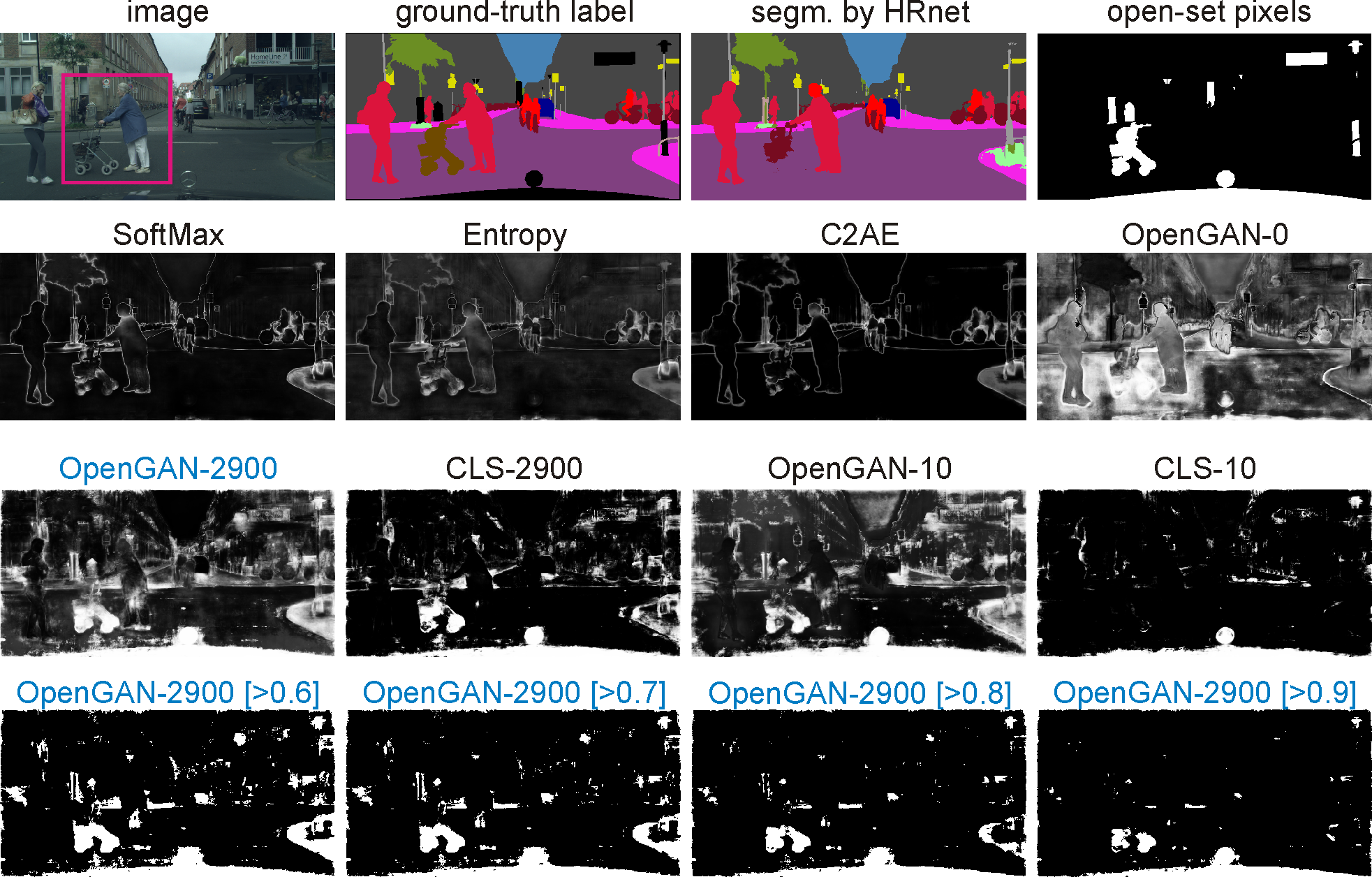}
\end{center}
\vspace*{-3mm}
\caption{\small
{\bf Qualitative results of a testing image from Cityscapes}.
[1$^{st}$ row] the input image, its per-pixel semantic labels,
the semantic segmentation result by HRnet and open-set pixels colored by white.
[2$^{nd}$ row] visual results as per-pixel scores of being classified as open-set pixel by SoftMax, Entropy, C2AE
and our OpenGAN-0$^{fea}$.
[3$^{rd}$ row] visual results by our OpenGAN$^{fea}$ and CLS,
trained with 2900 or 10 open training images, respectively.
[4$^{th}$ row] visual results by thresholding OpenGAN-2900 with 0.6, 0.7. 0.8 and 0.9 respectively.
OpenGAN clearly captures most open-set pixels (cf. the white pixels in top-right open-set map).
}
%\vspace{-5mm}
\label{fig:qualitativeDemo_appendix_288}
\end{figure*}

\begin{figure*}[t]
\begin{center}
   \includegraphics[width=0.75\linewidth]{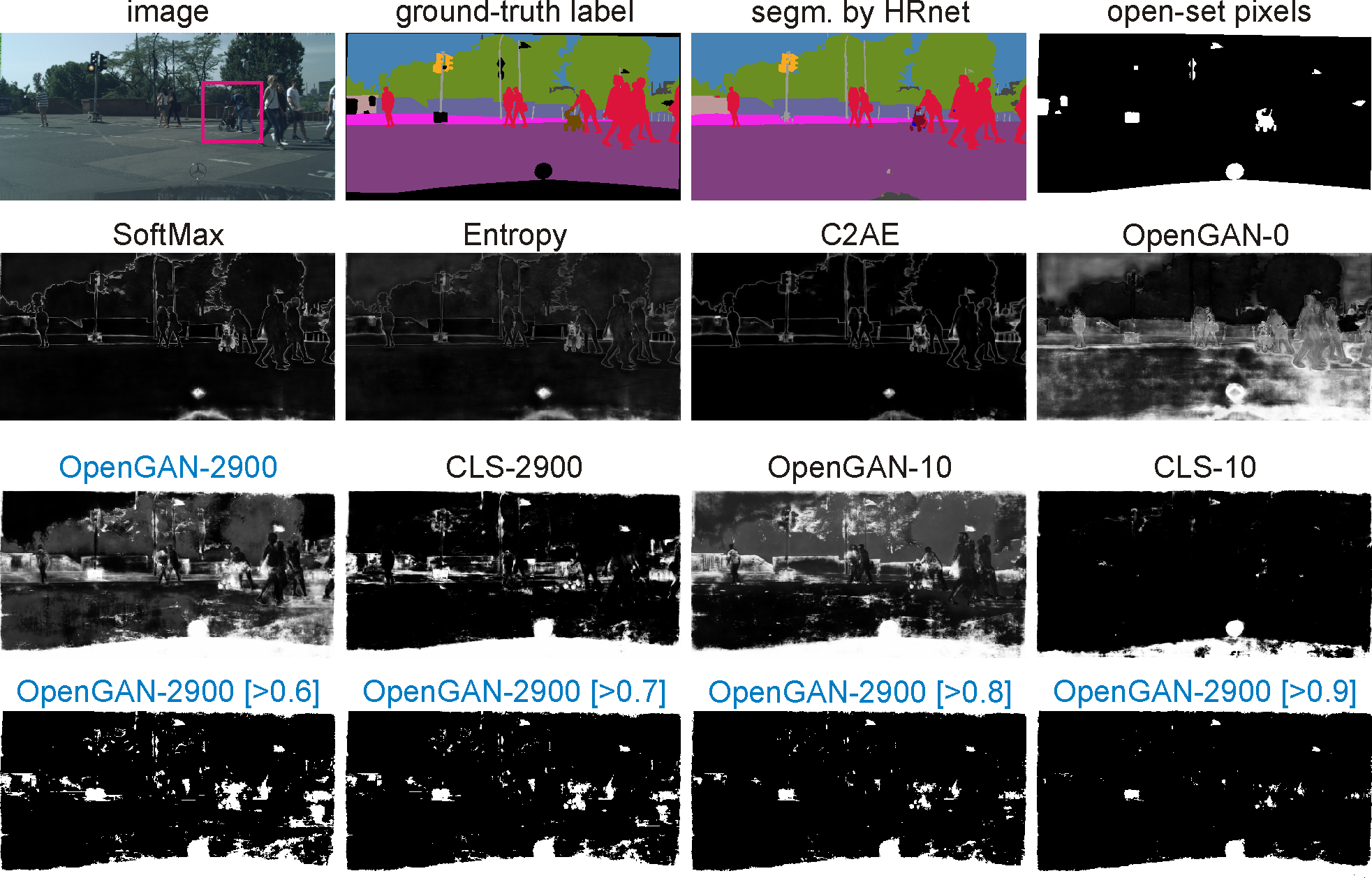}
\end{center}
\vspace*{-3mm}
\caption{\small
{\bf Qualitative results of a testing image from Cityscapes}.
[1$^{st}$ row] the input image, its per-pixel semantic labels,
the semantic segmentation result by HRnet and open-set pixels colored by white.
[2$^{nd}$ row] visual results as per-pixel scores of being classified as open-set pixel by SoftMax, Entropy, C2AE
and our OpenGAN-0$^{fea}$.
[3$^{rd}$ row] visual results by our OpenGAN$^{fea}$ and CLS,
trained with 2900 or 10 open training images, respectively.
[4$^{th}$ row] visual results by thresholding OpenGAN-2900 with 0.6, 0.7. 0.8 and 0.9 respectively.
OpenGAN clearly captures most open-set pixels (cf. the white pixels in top-right open-set map).
}
%\vspace{-5mm}
\label{fig:qualitativeDemo_appendix_193}
\end{figure*}

\begin{figure*}[t]
\begin{center}
   \includegraphics[width=0.75\linewidth]{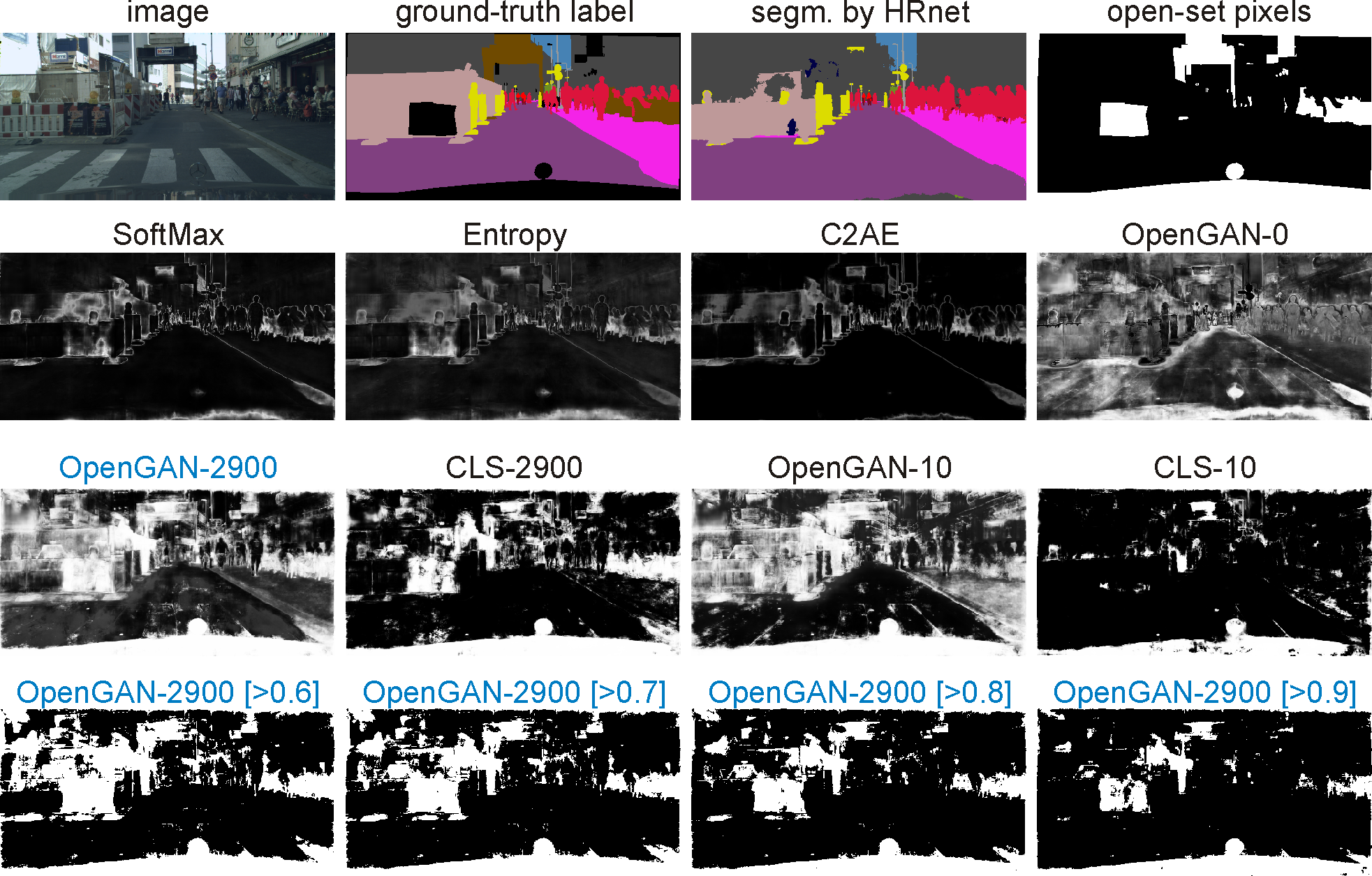}
\end{center}
\vspace*{-3mm}
\caption{\small
{\bf Qualitative results of a testing image from Cityscapes}.
[1$^{st}$ row] the input image, its per-pixel semantic labels,
the semantic segmentation result by HRnet and open-set pixels colored by white.
[2$^{nd}$ row] visual results as per-pixel scores of being classified as open-set pixel by SoftMax, Entropy, C2AE
and our OpenGAN-0$^{fea}$.
[3$^{rd}$ row] visual results by our OpenGAN$^{fea}$ and CLS,
trained with 2900 or 10 open training images, respectively.
[4$^{th}$ row] visual results by thresholding OpenGAN-2900 with 0.6, 0.7. 0.8 and 0.9 respectively.
OpenGAN clearly captures most open-set pixels (cf. the white pixels in top-right open-set map).
}
%\vspace{-5mm}
\label{fig:qualitativeDemo_appendix_125}
\end{figure*}

\begin{figure*}[t]
\begin{center}
   \includegraphics[width=0.75\linewidth]{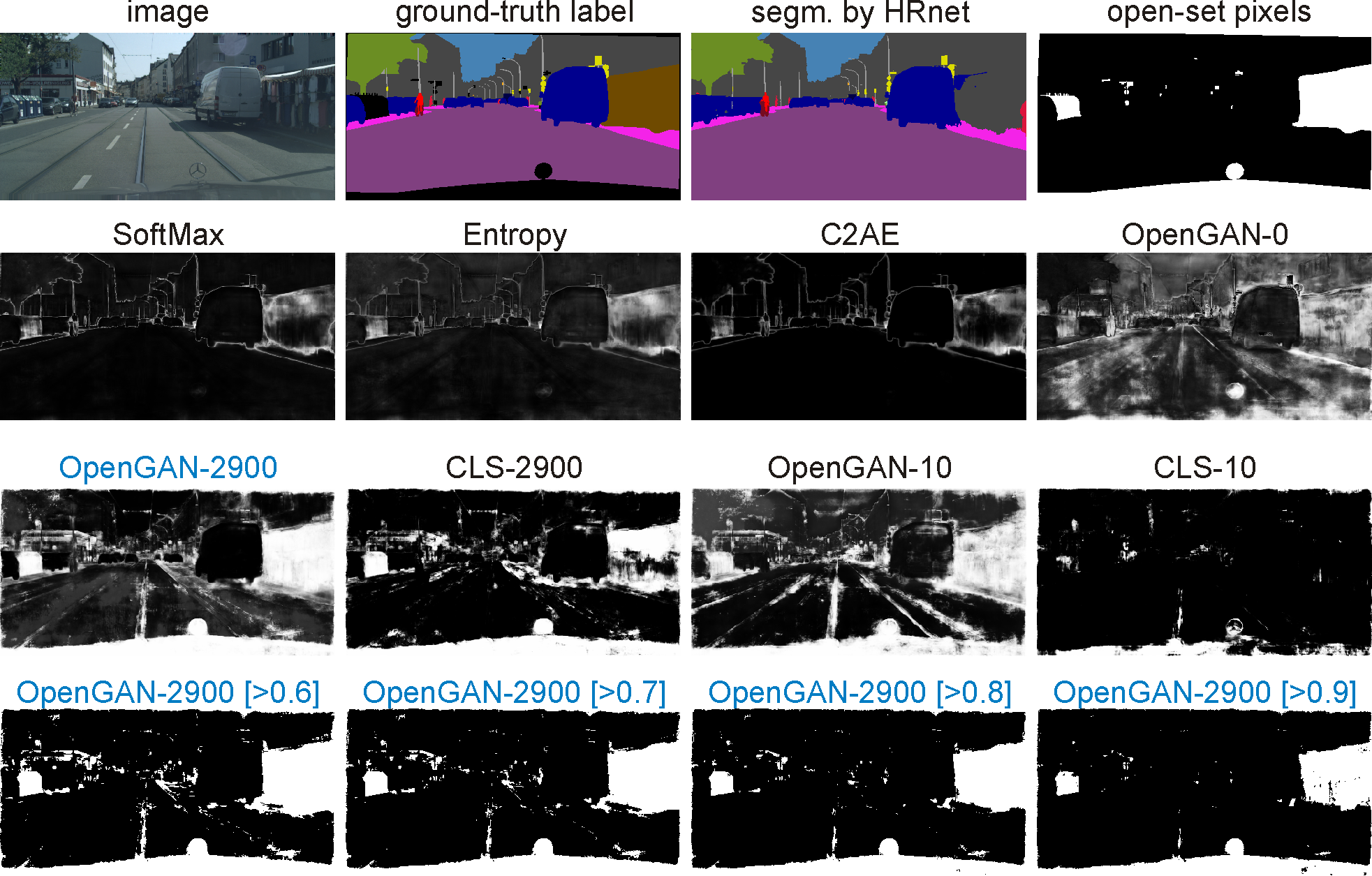}
\end{center}
\vspace*{-3mm}
\caption{\small
{\bf Qualitative results of a testing image from Cityscapes}.
[1$^{st}$ row] the input image, its per-pixel semantic labels,
the semantic segmentation result by HRnet and open-set pixels colored by white.
[2$^{nd}$ row] visual results as per-pixel scores of being classified as open-set pixel by SoftMax, Entropy, C2AE
and our OpenGAN-0$^{fea}$.
[3$^{rd}$ row] visual results by our OpenGAN$^{fea}$ and CLS,
trained with 2900 or 10 open training images, respectively.
[4$^{th}$ row] visual results by thresholding OpenGAN-2900 with 0.6, 0.7. 0.8 and 0.9 respectively.
OpenGAN clearly captures most open-set pixels (cf. the white pixels in top-right open-set map).
}
%\vspace{-5mm}
\label{fig:qualitativeDemo_appendix_201}
\end{figure*}

\begin{figure*}[t]
\begin{center}
   \includegraphics[width=0.75\linewidth]{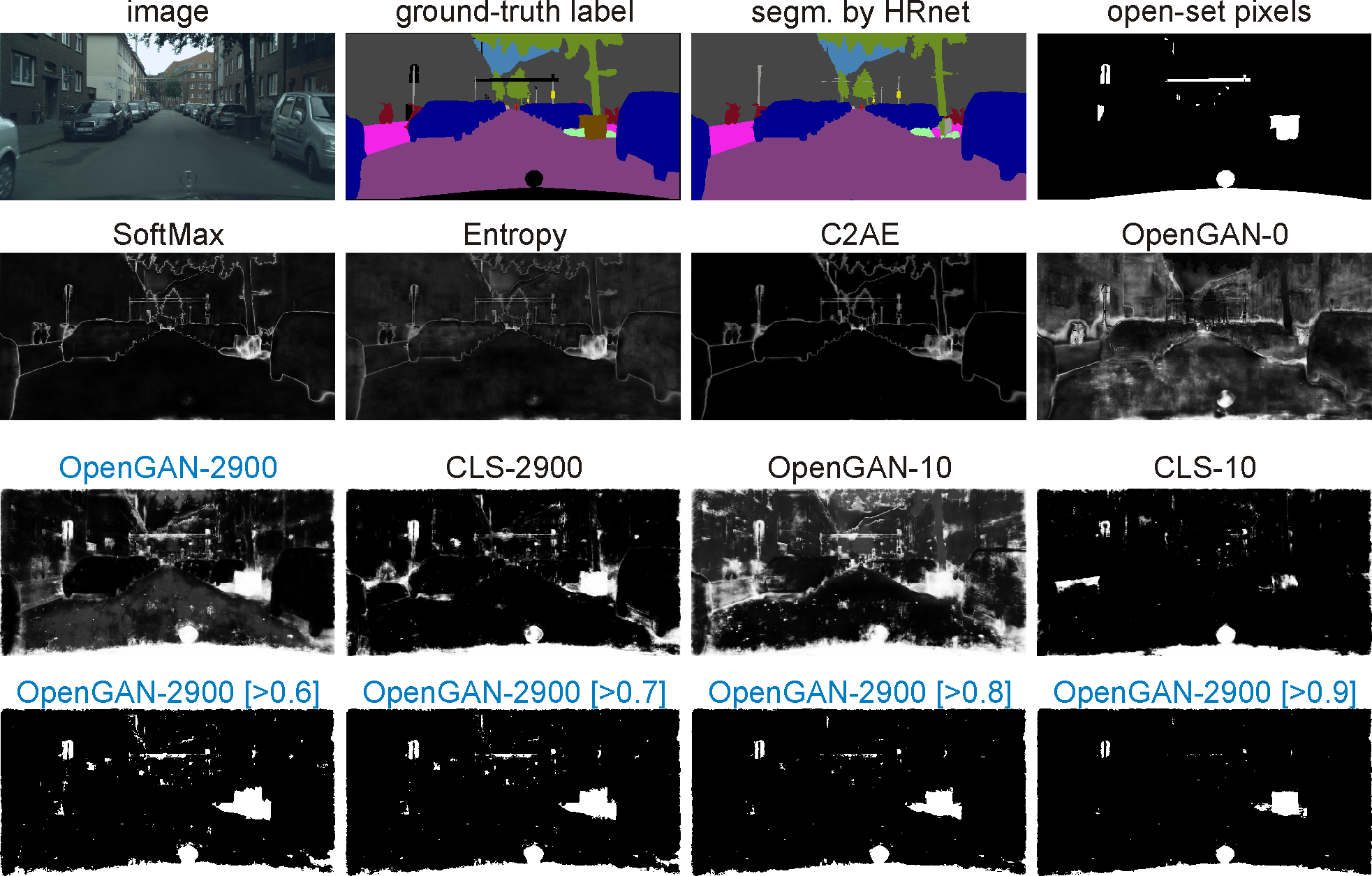}
\end{center}
\vspace*{-3mm}
\caption{\small
{\bf Qualitative results of a testing image from Cityscapes}.
[1$^{st}$ row] the input image, its per-pixel semantic labels,
the semantic segmentation result by HRnet and open-set pixels colored by white.
[2$^{nd}$ row] visual results as per-pixel scores of being classified as open-set pixel by SoftMax, Entropy, C2AE
and our OpenGAN-0$^{fea}$.
[3$^{rd}$ row] visual results by our OpenGAN$^{fea}$ and CLS,
trained with 2900 or 10 open training images, respectively.
[4$^{th}$ row] visual results by thresholding OpenGAN-2900 with 0.6, 0.7. 0.8 and 0.9 respectively.
OpenGAN clearly captures most open-set pixels (cf. the white pixels in top-right open-set map).
}
%\vspace{-5mm}
\label{fig:qualitativeDemo_appendix_422}
\end{figure*}

On the task of open-set semantic segmentation,
first, we show in Figure~\ref{fig:street-shop} that our testing set contains real open-set examples never-before-seen in training; please refer to the caption for details.
Then, we show more visual results in
figures from  \ref{fig:qualitativeDemo_appendix_136} through \ref{fig:qualitativeDemo_appendix_422}.
From these figures, we can see OpenGAN$^{fea}$  captures most open-set pixels, outperforming the other methods notably.

\section{Failure Cases and Limitations}
\label{sec:limitation}

As we use an discriminator as the open-set likelihood function,  straightforwardly, failure cases happen when the classification is not correct, as shown by the marked confidence scores in Figure~\ref{fig:viz_toy_example}, as well as the thresholded per-pixel predictions in figures from  \ref{fig:qualitativeDemo_appendix_136} through \ref{fig:qualitativeDemo_appendix_422}.
% \ref{fig:qualitativeDemo_appendix_136}, \ref{fig:qualitativeDemo_appendix_130},  \ref{fig:qualitativeDemo_appendix_193}, \ref{fig:qualitativeDemo_appendix_125}, \ref{fig:qualitativeDemo_appendix_201}, \ref{fig:qualitativeDemo_appendix_422}.

Hereby we point out other failure cases and limitations.
First, as we have explained in the main paper, the GAN-discriminator will eventually become incapable of discriminating closed-set and fake/open-set images due to the nature of GANs that strikes an equilibrium between the discriminator and generator. Although we empirically show superior performance by model selection on a validation set, there surely exists risks that the validation set is biased in an unknown way which could catastrophically hurt the final open-set recognition performance. This is also true even with outlier training examples.
Therefore, in the real open-world practitioners should be aware of such a bias, and exploit prior knowledge in constructing ``reliable'' training and validation sets in trainign OpenGANs.
Second, as we adopt adversarial training for OpenGANs, it is straightforward to ask if OpenGAN is robust to adversarial perturbations on the input images. We have not investigated this point yet, and we leave it as future work.

\end{document}